\documentclass{article}

\PassOptionsToPackage{numbers, compress}{natbib}
\usepackage[nonatbib,preprint]{neurips_2023}

\usepackage[utf8]{inputenc} % allow utf-8 input
\usepackage[T1]{fontenc}    % use 8-bit T1 fonts
\usepackage{hyperref}       % hyperlinks
\usepackage{url}            % simple URL typesetting
\usepackage{booktabs}       % professional-quality tables
\usepackage{amsfonts}       % blackboard math symbols
\usepackage{nicefrac}       % compact symbols for 1/2, etc.
\usepackage{microtype}      % microtypography
\usepackage{xcolor}         % colors
\usepackage{amsmath}
\usepackage{environ}
\usepackage{tikz}
\usepackage{setspace}
\usepackage{algorithm,algpseudocode}
\usetikzlibrary{shapes.geometric, arrows}
\usepackage{biblatex}
\usepackage{import}

\NewEnviron{Remarks}%
  {\vspace{0.5cm}
  \textbf{Remarks:}
  \begin{itemize}
      \BODY
  \end{itemize}
  }

\usepackage[normalem]{ulem}

\usepackage{setspace}
\usepackage{comment}
\usepackage{amsmath,amsthm,amssymb}
\usepackage{ulem}
\usepackage{microtype}
\DisableLigatures{}

\newcommand{\bs}{\boldsymbol}

\newcommand{\refeqp}[1]{Eq. (\ref{#1})}

\newcommand{\ee}{\end{equation}}
\newcommand{\be}{\begin{equation}}
\newcommand{\ec}{\end{center}}
\newcommand{\bc}{\begin{center}}
\newcommand{\eea}{\end{eqnarray}}
\newcommand{\bea}{\begin{eqnarray}}
\newcommand{\bd}{\begin{description}}
\newcommand{\ed}{\end{description}}
\newcommand{\bi}{\begin{itemize}}
\newcommand{\ei}{\end{itemize}}
\newcommand{\pa}{\partial}

\newcommand{\bx}{\bs{x}}

\newcommand{\by}{\bs{y}}

\newcommand{\bz}{\bs{z}}

\usepackage{commath}
\usepackage{mathtools}
\usepackage{hyperref}
\usepackage{color}
\usepackage{amsfonts}

\usepackage{diagbox}
\usepackage{bm}
\usepackage{booktabs}       % professional-quality tables
\usepackage{multirow}
\usepackage{subfigure}
\usepackage[percent]{overpic}
\usepackage{caption}
\usepackage{siunitx}

\newcommand{\true}{{\mathrm{true}}}

\newcommand{\ve}[1]{\boldsymbol{#1}}
\newcommand{\diff}[1]{\mathrm{d} #1}

\renewcommand{\u}[0]{U}
\newcommand{\bu}[0]{\ve \u}
\newcommand{\hu}[0]{\hat{\u}}
\newcommand{\hbu}[0]{\hat{\bu}}

\renewcommand{\by}[0]{\ve u}% alternative
\renewcommand{\bz}[0]{\ve u} 
\newcommand{\z}[0]{u}

\newcommand{\x}[0]{m}
\newcommand{\X}[0]{M}
\renewcommand{\bx}[0]{\ve \x}

\newcommand{\tbx}[0]{\tilde{\bx}}

\renewcommand{\chi}{\sigma}
\newcommand{\sig}[0]{\Sigma}
\newcommand{\bchi}[0]{\ve \sigma}

\newcommand{\bsig}[0]{\ve \Sigma}
\newcommand{\tsig}[0]{\tilde{\sig}}
\newcommand{\tbsig}[0]{\tilde{\bsig}}

\newcommand{\W}[0]{W}

\newcommand{\bW}[0]{\ve \W}

\newcommand{\rc}[0]{r^{(c)}}

\newcommand{\re}[0]{r^{(e)}}
\newcommand{\hre}[0]{\hat{r}^{(e)}}
\newcommand{\hrc}[0]{\hat{r}^{(c)}}

\newcommand{\bRc}[0]{\ve R^{(c)}}
\newcommand{\bRe}[0]{\ve R^{(e)}}
\newcommand{\hbRe}[0]{\hat{\ve R}^{(e)}}
\newcommand{\hbRc}[0]{\hat{\ve R}^{(c)}}

\newcommand{\lamc}[0]{\lambda^{(c)}}
\newcommand{\blamc}[0]{\ve \lambda^{(c)}}
\newcommand{\lame}[0]{\lambda^{(e)}}

\newcommand{\q}[0]{q_{\ve \xi}}
\newcommand{\qfull}[0]{q_{\ve \xi}(\bz, \bx, \bchi, \blamc)}
\newcommand{\s}[0]{\ve s}

\title{The Illusion of Fit: Spatially Resolved Assessment of Constitutive 
Model Validity in Elastography and Physics-Based Inverse Problems}

\author{%
  Vincent C. Scholz$^a$, P.S. Koutsourelakis$^{a,b}$\\
  $^a$ Technical University of Munich, Professorship of Data-driven Materials Modeling,  \\
  School of Engineering and Design, Boltzmannstr. 15, Garching, Germany \\
  $^b$ Munich Data Science Institute (MDSI - Core member), Garching, Germany \\
  \texttt{vincent.scholz@tum.de, p.s.koutsourelakis@tum.de} \\
}

\begin{document}

\maketitle

\begin{abstract}
Inferring the mechanical properties of soft tissues from measured 
deformations is a fundamental challenge in elastography and, more 
broadly, in physics-model-based inverse problems. A critical but 
rarely examined assumption underlying all existing approaches, i.e. 
direct, indirect, and learning-based alike, is that the assumed 
constitutive law correctly describes the material being imaged. When 
this assumption fails, inversion still produces parameter estimates 
that appear plausible, creating an \emph{illusion of fit}: 
practitioners obtain seemingly reasonable material property 
distributions with no indication that the constitutive model may be 
invalid in parts of the domain, a situation that can actively 
mislead clinical interpretation.

We propose a probabilistic framework that transforms constitutive 
model validity from an implicit assumption into an explicit, 
spatially resolved inference target. The key architectural departure 
from standard formulations is to treat the stress field as an 
independent latent variable rather than deriving it from displacements 
and material properties through the constitutive law. This enables a 
pointwise comparison between the stress required by mechanical 
equilibrium  and the 
stress predicted by the assumed constitutive model. Both sets of 
governing equations are incorporated into the probabilistic learning 
objective as virtual observables with separate precision 
hyperparameters: the conservation law precision is set a priori to 
a small value reflecting its undisputed validity, while the 
constitutive precision is inferred from the data under a 
sparsity-promoting prior. The resulting spatial distribution of 
constitutive precisions provides an uncertainty-aware map of where 
the assumed model is supported by the data and where it is not. 
Inference is carried out via stochastic variational inference, 
making the approach forward-model-free.

We validate the framework on synthetic harmonic elastography 
experiments using a brain slice geometry with a localized anisotropic 
inclusion embedded in an otherwise linear elastic domain. The 
inferred precision field correctly identifies the inclusion with a 
contrast of five orders of magnitude relative to the valid 
surrounding domain, robust across noise levels from 35 to 25 dB and 
under a four-fold reduction in observation density. A phantom 
experiment using real ultrasound-based displacement measurements from 
a material known to follow linear elastic behavior confirms that the 
method produces no false positive constitutive violations, while 
recovering the true stiffness contrast within the $99\%$ credibility 
interval. The framework is applicable to any PDE-constrained inverse 
problem in which conservation laws can be trusted but closure 
relations cannot.

\end{abstract}

% ================== MAIN CONTENT ==================

\section{Introduction}
Elastography 
is a technique used to infer the mechanical properties of soft tissues, aiding in disease 
diagnosis and treatment planning \cite{ophir1991elastography, doyley2012model, ormachea_elastography_2020}. Data acquisition typically involves applying static, 
harmonic, or transient excitations to tissue and measuring the resulting deformation using 
imaging modalities such as ultrasound \cite{ophir1991elastography}, magnetic resonance 
imaging \cite{muthupillai1995magnetic, sack_magnetic_2022}, or optical coherence 
tomography \cite{khalil2005tissue}. Recovering material parameters of the underlying tissue  from these measurements 
requires solving an inverse problem governed by the equations of solid mechanics, and a 
large body of methods has been developed to this end \cite{doyley2012model, 
fovargue_stiffness_2018}.

% ----------------------------------------------------------------------
% Direct and indirect methods
Reconstructing tissue properties from such measurements can be approached in two broad 
ways: direct and indirect methods \cite{mcgarry2011comparison}. Direct methods utilize 
the governing equations of solid mechanics as algebraic equations for the material 
parameters of interest, where the inferred strains and their derivatives appear as 
coefficients \cite{babaniyi_direct_2017, sumi1995estimation, manduca2001magnetic}. Their 
main appeal lies in computational efficiency and minimal dependence on boundary 
conditions. However, they rely on smoothed, post-processed displacement data 
\cite{manduca2003spatio}, which can distort fine-scale features and make results highly 
sensitive to noise, artifacts, and limited resolution. Indirect methods rely on the same 
governing equations but formulate elastography as a model-based optimization problem, 
minimizing the mismatch between observed and predicted displacements by repeatedly 
solving a forward model and usually its adjoint \cite{goenezen_linear_2012, 
oberai2003solution}. These approaches handle noisy and incomplete data more gracefully 
and accommodate complex material behavior, but at significantly higher computational 
cost. Both classes of methods, however, deliver only point estimates of material properties.

Quantifying uncertainty in these estimates, a critical need given 
that ambiguities in reconstructions carry direct diagnostic 
implications \cite{barbone_estimating_2011}, requires Bayesian 
formulations, which range from accurate but computationally intensive 
MCMC methods \cite{green2015bayesian, risholm2011probabilistic} to more
scalable, especially in time-sensitive clinical settings, variational approximations \cite{biehler_towards_2015, 
franck_multimodal_2017}.

% ----------------------------------------------------------------------
% Learning-based alternatives
Learning-based approaches have recently emerged as computationally attractive 
alternatives. Physics-informed neural networks (PINNs) \cite{raissi2019physics, 
kamali2023elasticity, haghighat2021physics, van2025enforcing, xu2023transfer} embed the 
governing equations directly into the training loss as residual penalties -- a philosophy 
of incorporating physical constraints into a learning objective that is shared, at an 
abstract level, with the present work. However, in PINNs, these constraints are enforced at inference time through strong residual penalties, which leads to difficulties with higher-order derivative computation and manual loss weight tuning; their reliance on continuous network approximations further results in poor handling of discontinuities common in elasticity \cite{van2025enforcing, raissi2019physics, zang2020weak}. Offline surrogate approaches \cite{patel_circumventing_2019}, including operator learning 
\cite{tripura_wavelet_2023} and diffusion-based methods \cite{dasgupta_conditional_2025},  train on labeled data pairs and achieve accelerated 
inference at test time, but require extensive and problem-specific training data, lack 
embedded physical guarantees during inference, and provide no indication of reliability 
when applied outside their training distribution, a critical concern in medical 
applications. AutoP methods \cite{hoerig_data-driven_2019, hoerig2021machine, 
newman2024improving} take a different route, recovering constitutive behavior directly 
from force-displacement data using neural networks without assuming a parametric 
constitutive form. While conceptually appealing, AutoP methods typically assume a single 
spatially uniform constitutive law and struggle with spatial variability and noisy data, 
limiting their applicability in heterogeneous tissue imaging.

% ----------------------------------------------------------------------
% The shared blind spot: constitutive model error
Despite their differences in methodology and computational philosophy, 
all of the approaches above share a critical and rarely examined 
assumption: that the governing equations,  and in particular the 
constitutive law relating stress to strain, correctly describe the 
tissue being imaged. While the conservation of linear momentum is 
grounded in fundamental physics and can be treated as reliable up to 
discretization error, constitutive laws are phenomenological \cite{holzapfel2017similarities}: they are selected on the basis of physical intuition and mathematical convenience rather than derived from first principles, and are rarely challenged once adopted. Tissue behavior is heterogeneous and pathology-dependent, 
and the assumed model class may break down in precisely the regions of 
greatest diagnostic interest. When the constitutive model is wrong, 
inversion still produces an estimate of material parameters, but 
those estimates are artifacts of the model mismatch rather than 
reflections of true tissue properties. As we demonstrate in this 
paper, such estimates can actively mislead clinical interpretation: an 
invalid constitutive assumption can produce a spuriously soft or stiff 
region that mimics a clinically meaningful finding, when in reality no 
valid material property estimate can be made under the assumed model. 
This risk is not limited to any particular inversion strategy; it 
affects direct and indirect methods, Bayesian and deterministic 
formulations, and learning-based approaches equally. Nor is it unique 
to elastography: in the vast majority of physics-model-based inverse 
problems, the governing equations comprise conservation laws (which 
are grounded in fundamental physics and can be treated as reliable) 
and constitutive or closure relations, which are phenomenological 
and whose validity is routinely assumed but rarely verified 
\cite{kaipio2006statistical,bonnet2005inverse}. Whether the 
application is subsurface geomechanics, cardiovascular mechanics 
\cite{biehler_towards_2015}, or structural health monitoring, the 
same risk arises: an invalid closure assumption produces estimates 
that appear numerically consistent yet carry no reliable physical 
meaning. Elastography represents a particularly consequential 
instance of this general problem, because the estimates derived from 
an invalid material model directly inform clinical decisions about 
disease diagnosis and treatment planning.

% ----------------------------------------------------------------------
% Existing approaches to model error and their limitations
Several strategies have been proposed for addressing model-form uncertainty in inverse 
problems. The Kennedy-O'Hagan (KOH) framework \cite{kennedy2001bayesian} introduces an 
additive discrepancy term, typically a Gaussian process, to correct model predictions 
\cite{Bayarri2009, Brynjarsdottir2014, Plumlee2017, Leoni2024}. While widely used in 
engineering applications, KOH has not been applied in elastography, in part because the 
discrepancy term absorbs prediction errors implicitly rather than attributing them to 
constitutive model inadequacy, leaving the physical source of model error unquantified. 
Moreover, identifiability issues arise because the discrepancy can obscure errors in the 
inferred material parameters, and the learned correction need not satisfy mechanical 
equilibrium, potentially producing physically inconsistent reconstructions. Parameter 
Uncertainty Inflation (PUI) \cite{pernot2017critical, sargsyan2015statistical} treats 
model error as parameter misspecification by inflating parameter variances, but in 
elastography,  where structural model errors and parameter effects are tightly coupled, this conflation leads to confounding and non-uniqueness 
\cite{pernot2017critical, arcones2024embedded}. Model selection via Bayes factors 
\cite{kass1995bayes} offers a principled comparison between competing constitutive model 
classes, but requires repeated inversion and cannot identify spatially localized failures 
\emph{within} a single model class. Data-driven constitutive discovery 
\cite{fuhg_review_2024, flaschel_automated_2023-1, wang2021inference, thakolkaran2022nn, 
joshi2022bayesian} offers greater expressiveness, but methods that operate without 
stress measurements (as required in elastography) typically assume a single global 
law and can produce an excellent fit to data regardless of whether the recovered model is 
physically meaningful \cite{thakolkaran2022nn, joshi2022bayesian, wang2021inference, 
man2011neural}. Increasing model expressiveness does not inherently solve the problem of 
model validity, but in fact it may simply obscure it \cite{agrawal_probabilistic_2024}. Crucially, none of these approaches provide 
a spatially resolved, uncertainty-aware map of \emph{where} a given constitutive 
assumption is supported by the data and where it is not.

% ----------------------------------------------------------------------
% Our method
We propose a probabilistic framework that addresses this gap directly. The key 
distinction we draw on is between \emph{reliable} governing equations, i.e. conservation 
laws grounded in fundamental physics,  and \emph{unreliable} ones, ie. constitutive 
relations that are phenomenological and potentially invalid \cite{feissel_modified_2007}. 
This distinction, introduced in the context of the modified error in constitutive 
equations (MECE) \cite{banerjee_large_2013,koutsourelakis2012novel}, is here embedded in a fully Bayesian 
probabilistic framework and combined with the virtual observable formulation  of 
\cite{kaltenbach2020incorporating, scholz2025weak, %koutsourelakis2012novel, 
%bruder2018beyond,
vadeboncoeur2023fully}. Both the conservation law and the constitutive 
relation are incorporated into the probabilistic learning objective as soft constraints, 
via separate sets of weighted residuals with separate precision hyperparameters governing 
the degree to which each is enforced. Critically, the stress field is treated as an 
independent latent variable rather than being derived from displacements and material 
parameters through the constitutive law. This enables a direct pointwise comparison 
between the stress required by mechanical equilibrium and the stress predicted by the 
assumed constitutive model. It is the discrepancy between these two that generates 
the constitutive validity signal. A sparsity-promoting Automatic Relevance Determination (ARD) prior on the constitutive 
precision hyperparameters then yields a spatially resolved, uncertainty-quantified map 
of where the assumed constitutive law is consistent with the data and where it fails. 
Inference is carried out via stochastic variational inference \cite{hoffman2013stochastic}, 
making the approach computationally tractable without requiring explicit forward model 
solves.

The contributions of this paper are as follows:

\begin{itemize}

    \item \textbf{Spatially resolved constitutive validity assessment:} The framework 
    produces a pointwise map of where the assumed constitutive law is supported by the 
    data, enabling targeted model refinement and providing an explicit warning when 
    inferred material properties should not be trusted diagnostically. To our knowledge, 
    this is the first method in elastography to provide such a map in a fully 
    probabilistic setting.

    \item \textbf{Robustness to incomplete data and uncertain boundary conditions:} By 
    treating all state variables, i.e. displacements, stresses, and material properties, 
    as random variables and enforcing governing equations probabilistically, the 
    framework tolerates realistic levels of measurement noise, sparse observations, and 
    missing displacement components without requiring knowledge of boundary conditions 
    \cite{scholz2025weak}.

    \item \textbf{Scalable forward-model-free inference:} Governing equations enter the 
    inference objective directly as weighted residuals rather than through a black-box 
    forward solver, eliminating the need for repeated forward solves and enabling 
    scalable parameter estimation and uncertainty quantification via stochastic 
    variational inference.

\end{itemize}

The remainder of this paper is organized as follows. Section~2 introduces the forward 
and Bayesian inverse problem formulations for elastography and presents the proposed 
methodology in detail, including the derivation of the evidence lower bound and the form 
of the approximate posterior. Section~3 validates the framework through synthetic 
experiments on a brain slice geometry and experimental phantom data. Section~4 concludes 
with a summary and outlook.

\section{Methodology} \label{sec:method}
\begin{figure}
    \centering
    \includegraphics{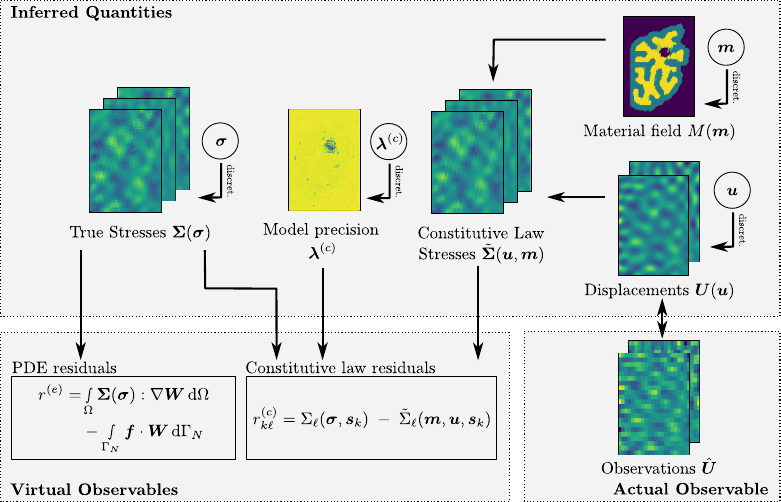}
       \caption{\textbf{Schematic overview of the proposed methodology 
(Section~\ref{sec:method}).} The framework simultaneously infers displacements $\bu$, material 
properties $\X$, and stress fields $\bsig$ from three data sources: 
noisy displacement observations $\hbu$, the weighted residuals of the 
conservation law (virtual likelihood $p(\hbRe|\bchi)$ in \refeqp{eq:likere}), and the 
residuals comparing $\bsig$ against the stresses $\tbsig$ predicted 
by the assumed constitutive law (virtual likelihood 
$p(\hbRc|\bz,\bx,\bchi,\blamc)$ in \refeqp{eq:likerc}). The stress field $\bsig$ is 
treated as an independent latent variable rather than being derived 
from $\bu$ and $\X$ through the constitutive law. This enables a 
pointwise comparison between the stress required by mechanical 
equilibrium and the stress the constitutive law predicts: their 
discrepancy, captured by the precision field $\blamc$, provides a 
spatially resolved map of constitutive model validity.
}
    \label{fig:scheme}
\end{figure}

We first introduce the key components of model-based elastography (section \ref{sec:problem}) and subsequently present the core aspects of the proposed formulation (section \ref{sec:main}). An overview is presented in Figure~\ref{fig:scheme}. Section \ref{sec:inference} presents both the computational methods for the sought material property parameter identification and model error quantification, and the corresponding pseudo-algorithm.

\subsection{Problem definition}
\label{sec:problem}
The goal of elastography is to identify material properties (e.g., elastic modulus) and their spatial variability, which we represent with a field $\X(\bs{s})$ where $\bs{s}$ denotes the spatial coordinates. 
This appears in the context of  a, typically local, constitutive law of solid mechanics (i.e.,  a stress-strain relation) of the form:
\begin{equation}
    \bsig(\s) = \bsig(\bs{\epsilon} (\s, \bu); ~\X(\s)), \label{eqn:constitutive}
\end{equation}
where $\bsig$ is the Cauchy stress tensor as a function of space $\s$, $\bu$ is the displacement field, $\bs{\epsilon}=\frac{1}{2}(\nabla \bu+(\nabla \bu)^T)$ is the infinitesimal strain tensor and $\X$ is the aforementioned sought,  material property field. %, linking those quantities with the stresses $\bsig$. 
Such constitutive laws (or, more generally, closure models) do not arise from undisputed physical principles, are phenomenological,  and constitute the model's {\em unreliable} component \cite{feissel_modified_2007}.

In the context of  indirect elastographic techniques, the aforementioned constitutive law is  supplemented by the conservation of linear momentum which in the  simplest, time-independent setting, corresponds to the following  Partial Differential Equation (PDE)    and accompanying boundary conditions:
\begin{align}
    \nabla \cdot \bsig(\s) &= 0, \quad \s \in \Omega \subset \mathbb{R}^d, \label{eqn:PDE} \\
    \bu(\s) &= \ve g, \quad \s \in \partial\Omega_\u \subseteq \pa \Omega , \ \mathrm{and}
    \\
    \bsig \cdot \ve n &= \ve f , \quad \s \in \pa \Omega_{\bsig}  =\pa \Omega  \setminus \pa \Omega_\u,
\end{align}
where $\ve n$ is the outward normal vector on the boundary $\pa \Omega$, and $\ve g$ and $\ve f$ are the prescribed Dirichlet boundary $\pa \Omega_\u$ and Neumann boundary $\pa \Omega_{\bsig}$, respectively. We note that boundary conditions might be partially or completely unavailable in some cases, a scenario we have addressed in previous work \cite{scholz2025weak}.  The  validity of the PDE above, as with all conservation laws (e.g.,  conservation of mass, energy),    is  undisputed. If any model errors are introduced through \refeqp{eqn:PDE}, then this will be solely due to the discretization that might be applied, e.g., when using FEM to solve the forward problem. 
As such, we refer to it as a {\em reliable} component of the model. 

To solve this problem, all fields must be represented in a finite-dimensional form. We adopt a finite-element-type discretization of the material field $\X$, the displacement field $\bu$, and the dependent state variables (stress field $\bsig$). Alternative representations, including neural-network-based approaches, are discussed in \ref{app:latent}. The coefficients $\bx = \{ \x_i \}_{i=0}^{d_{\bx}}$, $\bz = \{ \z_i \}_{i=0}^{d_{\bz}}$ and $\bchi = \{ \chi_i \}_{i=0}^{d_{\bchi}}$ represent the discretized versions of each field, respectively. Throughout, we assume the spatial resolution to be sufficiently fine that discretization errors are negligible (relative to other sources of uncertainty).

The aforementioned equations define a {\em forward model} with which one can  calculate displacements (output) $\bu$ from a given material property field (input) $\X$.   % plus boundary conditions 
 We denote abstractly the implied map as $\bu(\X)$.  This forward model is repeatedly solved by traditional, indirect elastographic techniques so that the fit to the observables is maximized. The latter consist of  noisy displacements $\hbu = \{ \hu_i \}_{i=1}^{N_{\hbu}}$ at locations $\hat{\s}=\{ \hat{s}_i \}_{i=1}^{N_{\hbu}}$, which (in the simplest case) are  assumed to   relate to the output/displacements of the forward problem as:
\begin{equation}
    \hbu(\hat{\s}) = \bu(\X, \hat{\s}) + \tau^{-0.5} \boldsymbol{\eta}, \quad \mathrm{with} \quad  \boldsymbol{\eta} \sim \mathcal{N}(\ve 0, \ve 1). \label{eqn:obs}
\end{equation}
The additive term accounts for the stochastic observation noise, which is typically assumed to be normally distributed with variance $\tau^{-1}$. The equation above defines the likelihood $p(\hbu| \X)$, which quantifies the probability of observing $\hbu$ given a material field $\X$ and assuming the model above is valid.  
 Combined  with a prior $p(\X)$,  it yields the posterior distribution:
\begin{equation}
    \begin{array}{ll}
        p(\X | \hbu) &  \propto p(\hbu| \X) ~p(\X) \\
         & \propto \prod_{i=1}^{N_{\hbu}} \mathcal{N} (\hu_i~|~ \u_i(\X), ~\tau^{-1}) ~~p(\X).
    \end{array}
    \label{eqn:posterior_unnormed} 
\end{equation}
The posterior probabilistically quantifies the plausibility of various material property fields $\X$ and   is  the sought quantity in the Bayesian formulation of model-based elastography.  Deterministic formulations can be considered as special cases where point-estimates, such as e.g.\ the MAP (Maximum A Posteriori) estimate, are sought \cite{doyley2012model}. 
In both frameworks, estimates of $\X$ that best fit the observed data can always be obtained—even if the underlying model, specifically the constitutive law in \refeqp{eqn:constitutive}, is invalid. However, when the model is incorrect, the inferred material properties lose their diagnostic significance and may even mislead clinical interpretation, potentially resulting in harmful or erroneous conclusions.

\noindent \textbf{Remark:} 
Our discussion is restricted to \textit{small-strain elasticity} for clarity and ease of presentation. However, the formulation can be directly extended to \textit{finite-strain elasticity}, where the Cauchy stress tensor \( \bsig \) is replaced by either the \textit{first Piola-Kirchhoff stress tensor} \( \bs{P} \) or the \textit{second Piola-Kirchhoff stress tensor} \( \bs{S} \), and the \textit{infinitesimal strain tensor} \( \bs{\epsilon} \) is substituted with measures appropriate for large deformations, such as the \textit{deformation gradient} \( \bs{F} \) or the \textit{Green-Lagrange strain tensor} \( \bs{E} \), within the constitutive relation (\refeqp{eqn:constitutive}). Extensions to time-harmonic settings \cite{zhang_solution_2012} are straightforward and  are discussed in the  numerical illustrations (section \ref{sec:results}).

\subsection{Main idea}
\label{sec:main}

\begin{figure}
    \centering
    \tikzstyle{observe} = [rectangle, 
    minimum width=0.6cm, 
    minimum height=0.6cm, 
    text centered, 
    text width=0.6cm, 
    draw=black, 
    fill=gray!30]
    
    \tikzstyle{latent} = [circle, 
    minimum width=0.6cm, 
    minimum height=0.6cm, 
    text centered, 
    draw=black]
    
    \tikzstyle{colorless} = [rectangle, 
    minimum width=0.6cm, 
    minimum height=0.6cm, 
    text centered, 
    text width=0.6cm, 
    draw=black]
    
    \tikzstyle{arrow} = [thick,->,>=stealth]
    
    \begin{tikzpicture}[node distance=1.5cm]
    % Original latent variables
    \node (x) [latent] {$\bx$};
    \node (z) [latent, below  of=x] {$\bz$};
    \node (chi) [latent, above of=x] {$\bchi$};
    
    % New colorless boxes layer with increased horizontal spacing
    \node (bsig) [colorless, right of=chi, node distance=3cm] {$\bsig$};
    \node (M) [colorless, right of=x, node distance=3cm] {$\X$};
    \node (U) [colorless, right of=z, node distance=3cm] {$\bu$};
    
    % Intermediate layer with increased horizontal spacing
    \node (tbsig) [colorless, right of=M, node distance=2.75cm] {$\tbsig$};
    
    % Observed variables
    \node (re) [observe, right of=tbsig] {$\hat{\bs{R}}^{(c)}$};
    \node (rc) [observe, above of=re] {$\hat{\bs{R}}^{(e)}$};
    \node (u) [observe, below of=re] {$\hbu$};
    \node (lame) [latent, right of=re] {$\blamc$};
    
    % Horizontal arrows from circles to colorless layer with "discretization" labels
    \draw [arrow] (chi) -- (bsig) node[midway,above,font=\scriptsize,sloped] {discretization};
    \draw [arrow] (x) -- (M) node[midway,above,font=\scriptsize,sloped] {discretization};
    \draw [arrow] (z) -- (U) node[midway,above,font=\scriptsize,sloped] {discretization};
    
    % Arrows from colorless layer to intermediate layer with "const. law" labels
    \draw [arrow] (U) -- (tbsig) node[midway,above,font=\scriptsize,sloped] {const. law};
    \draw [arrow] (M) -- (tbsig) node[midway,above,font=\scriptsize,sloped] {const. law};
    
    % Arrows from colorless layer to observables
    \draw [arrow] (bsig) -- (rc);
    \draw [arrow] (bsig) -- (re);
    \draw [arrow] (U) -- (u);
    
    % Arrow from intermediate layer to observables
    \draw [arrow] (tbsig) -- (re);
    
    % Lambda arrows remain as is
    \draw [arrow] (lame) -- (re);
    
    \end{tikzpicture}
    \caption{Probabilistic graphical model for the presented model. Circles represent latent (unobserved) variables, shaded rectangles represent (actually or virtually) observed variables, and non-shaded rectangles represent deterministically computed variables. Arrows indicate conditional dependencies between variables. All variables and their meanings are detailed in the following sections, with hyperparameters omitted for clarity.}
    \label{fig:chart}
\end{figure}

The key architectural departure from standard formulations lies in 
treating the stress field $\bsig$ as an independent latent variable, 
rather than computing it deterministically from the displacement field 
$\bu$ and material properties $\X$ via the constitutive law 
\eqref{eqn:constitutive}. In standard indirect methods, stress is 
always derived through the constitutive law, which means any 
discrepancy between the assumed model and the true tissue behavior is 
silently absorbed into the inferred material properties. 
By contrast, 
treating $\bsig$ as an independent variable allows us to separately 
ask: what stress field is required by mechanical equilibrium, and what 
stress field does the assumed constitutive law predict given the 
inferred displacements and material properties? The discrepancy between 
these two quantities is the signal from which constitutive model 
validity is locally assessed. We formalize this through two separate 
sets of residuals, i.e. one for each governing equation and thus two 
corresponding virtual likelihoods, each with its own precision 
hyperparameter governing the degree to which the corresponding equation 
is enforced. This extends the framework of \cite{scholz2025weak}, 
which incorporated only the conservation law as a virtual observable, 
by additionally treating the constitutive law as a soft constraint 
whose local validity is inferred from the data. The resulting 
probabilistic graphical model is shown in Figure~\ref{fig:chart}.

In particular, we  make use of the weighted residuals (based on the weak form of the PDE in Eq. \eqref{eqn:PDE}) for the conservation law 
\begin{equation}
    \re_{j}(\bchi) = \frac{1}{\int_\Omega \norm{\bW^{(j)}}_1 \diff\Omega}\left( \int_{\Omega} \bsig(\bchi) \!:\! \nabla \bW^{(j)} \diff\Omega - \int_{\Gamma_N}  \boldsymbol{f} \cdot \bW^{(j)} \diff\Gamma_N \right), \label{eqn:rc} %- \int_{\Omega} \rho b \cdot w \diff\Omega 
\end{equation}
where each residual differs in the selection of the (vector-valued) weight function $\bW^{(j)}$ \cite{finlayson_method_1972}. We note that in our formulation, the weighted residuals serve as data sources rather than as a means to derive a conventional discretized system of equations. The more residuals (i.e., weighting functions) are considered and incorporated, the more information about the corresponding PDE is encoded.  We normalize the residuals as shown above to ensure invariance under a uniform scaling of the weighting functions.

We consider a second set of residuals of collocation-type at points with coordinates $\s_k$ for each independent stress tensor component $N_\ell$ with index $\ell$ (leveraging Voigt notation):
\begin{equation}
    \rc_{k \ell}(\bx, \bz,\bchi) = \sig_\ell(\bchi, \s_k) - \tsig_\ell(\bx, \bz, \s_k) 
\end{equation}
% \psk{u had double indices for stress in the equation just above!}
which quantify the discrepancy between the actual (unknown) 
stresses $\bsig$ (expressed with respect to $\bchi$ as in  Eq. \eqref{eqn:sfield}) and the stresses $\tbsig$ computed from the assumed and potentially erroneous constitutive law in Eq. \eqref{eqn:constitutive}. 
 We note that the proposed method does not rely on the type of residuals employed but rather on considering residuals for each governing equation separately. As such, one could use other combinations of (weighted) residuals or norms.

The central device we employ is to treat the value of each residual 
as a \emph{virtual observable} \cite{kaltenbach2020incorporating, 
scholz2025weak}: just as actual displacement measurements are assumed 
to relate to the true displacement field through a noise model, we 
assume that each residual has been ``observed'' to take the value 
zero, subject to a precision parameter that controls how strictly this 
is enforced. This allows governing equations to enter the probabilistic 
learning objective as soft constraints rather than hard ones, with the 
degree of enforcement controlled by interpretable hyperparameters 
rather than fixed penalty weights. Similar to the actual observables  in Eq. \eqref{eqn:obs},  we postulate relations of the form:

\begin{align}
    0=\hre_{j} &= \re_j (\bchi)+\left(\lame \right)^{-1}\eta_j,  &&\ \mathrm{with} \ \eta_j \sim \mathcal{N}(0,1), \quad \mathrm{and} \\
    0=\hrc_{k \ell} &= \rc_{k \ell}(\bz,\bx, \bchi) + \left(\lamc_{k \ell} \right)^{-1}\eta_{k \ell}, &&\ \mathrm{with} \ \eta_{k \ell} \sim \mathcal{N}(0,1),
\end{align}
where the hyperparameters $\lame$ and $\bs{\lamc}=\{\lamc_{k \ell}\}_{k=1, \ell=1}^{N_c, N_\ell}$ express the error precisions.
These equations give rise to {\em virtual} likelihoods of the forms
\be
   % \phantom{p(\hbRc | \bz,\bx, \bchi, \blamc)}
    \begin{array}{ll}
        \mathllap{p(\hbRe | \bchi)} & =\prod_{j=1}^{N_e} p(\hre_j=0 | ~\bchi, \lame) \\
        & \propto \prod_{j=1}^{N_e} \sqrt{\lame}  \exp\left(  - \frac{\lame}{2} ~\left(\re_j(\bchi) \right)^2\right), 
        \end{array}
        \label{eq:likere}
    \ee
 and
\be
\begin{array}{ll}
p(\hbRc | \bz,\bx, \bchi, \blamc) 
   & = \prod_{k=1}^{N_c}\prod_{\ell=1}^{N_\ell} p(\hrc_{k\ell}=0 | \bz,\bx,\bchi,\blamc)\\
   & \propto \prod_{k=1}^{N_c}\prod_{\ell=1}^{N_\ell} \sqrt{\lamc_{k\ell}}\exp\!\left(-\tfrac{\lamc_{k\ell}}{2}\left(\rc_{k\ell}(\bz,\bx,\bchi)\right)^2\right),
\end{array}
\label{eq:likerc}
\ee
where the two sets of virtual observables are denoted summarily with $\hbRe=\{ \hre_{j}=0 \}_{j=1}^{N_e}$ and $\hbRc=\{ \hrc_{k \ell}=0 \}_{k=1, \ell=1}^{N_c, N_\ell}$.
We note that the hyper-parameters  $\{\lame, \blamc\}>0$ control the likelihood of the residuals which deviate from $0$, i.e. the value they would attain for  solution-tuples $\{ \bz,\bx, \bchi \}$ (or equivalently $\X$, $\bsig$ and $\bu$) that satisfy the governing equations. 
Given the undisputed validity of the conservation law (up to discretization errors), one can a priori set  $(\lame)^{-1}$ to very small values,  similar to the  tolerance of a deterministic iterative solver.

In contrast, and given that a-priori the validity of the constitutive law is not guaranteed,  $\blamc$  would need to be inferred.
If one were to fix $\blamc$ to prescribed values, the problem would 
be severely non-identifiable: in the extreme case $\lamc_{k\ell} = 0$ 
at all collocation points, the constitutive law is completely ignored 
and the inference reduces to finding any combination of $\bu$ and 
$\bsig$ that satisfies equilibrium, i.e. a massively underdetermined 
problem with no unique solution for $\X$. Conversely, if 
$\lamc_{k\ell} \to \infty$ everywhere, the constitutive law is 
enforced as a hard constraint and the framework collapses to a 
standard inverse problem with no capacity to detect model error. 
Neither extreme is desirable; what is needed is a data-driven 
mechanism that enforces the constitutive law where it is consistent 
with the observed data and relaxes it where it is not. We achieve 
this by \emph{inferring} $\blamc$ from the data under a principle of 
parsimony: among all combinations of $\X$, $\bu$, $\bsig$, and 
$\blamc$ that are consistent with the observations and the 
conservation law, we prefer those that satisfy the constitutive law 
at as many collocation points as possible, i.e., those for which 
the number of non-zero residuals $\rc_{k\ell}$ is minimized.
In other words, solutions  (i.e., combinations of $\X,\bu,\bsig$) that satisfy the constitutive law as much as possible or in as many collocation points as possible, are preferred over others.
To induce such a behavior, we use a sparsity-promoting prior for the hyperparameters $\blamc$, i.e., a hyperprior which promotes solutions in which $\rc_{k \ell}$ are zero or near zero. In particular, we employ the Automatic Relevance Determination (ARD) prior \cite{mackay_bayesian_1996,bishop_variational_2000} according to which:
\be
p(\blamc)=\prod_{k=1}^{N_c} \prod_{\ell=1}^{N_\ell} p(\lamc_{k \ell})=\prod_{k=1}^{N_c} \prod_{\ell=1}^{N_\ell} \mathrm{Gamma}(\lamc_{k \ell}|~a_0,b_0)
\label{eqn:prior_lamc}
\ee
where very small values $a_0=b_0=10^{-8}$ were used for the hyperparameters \cite{bishop_variational_2000}. This is preferred here over other sparsity-inducing  alternatives such as the spike-and-slab \cite{ishwaran_spike_2005} or the horseshoe \cite{carvalho_handling_2009} hyperpriors, due to its simplicity and the absence of hyperparameters requiring fine-tuning.
We note in \refeqp{eq:likerc} that a non-zero residual $\rc_{k \ell} \ne 0$ is penalized more (i.e., it has a lower (virtual) likelihood), the larger the corresponding $\lamc_{k \ell}$ is, and vice versa. Physically, $\lamc_{k \ell}$ reflects how well the stresses predicted by the constitutive law in Eq. \eqref{eqn:constitutive} (calculated from the local displacements and material field), are consistent with the governing equations in Eq. \eqref{eqn:PDE}: large values indicate a good local fit, while values near zero signal that the assumed constitutive law cannot produce stresses compatible with the trusted physics encoded in the PDE. 

The aforementioned {\em virtual} likelihoods are combined with the \textit{actual} likelihood of the observables $\hbu$ as in \refeqp{eqn:obs}, i.e.:
\begin{equation}
    \begin{array}{ll}
p(\hbu|\bz)  & =\prod_{i=1}^{N_{\hbu}} \mathcal{N} (\hu_i~|~ \u_i( \bz), ~\tau^{-1}) \\
& \propto \prod_{i=1}^{N_{\hbu}} \sqrt{ \tau}  e^{-\frac{\tau}{2} (\hu_i-\u_i( \bz,\hat{s}_i))^2},
\end{array}
\label{actual_like}
\end{equation}
in order to yield the joint posterior over all unknowns:

{\small
\begin{equation}
    p\left( \bx, \bz, \bchi, \blamc | \hbRc, \hbRe, \hbu \right) = \frac{p \left( \hbu|\bz \right) p\left(\hbRe | \bchi\right) p\left( \hbRc | \bz,\bx, \bchi, \blamc\right) p\left( \bx, \bz, \bchi \right) p\left(\blamc\right) }{p \left( \hbRc, \hbRe, \hbu \right) }.
    \label{eq:jposterior}
\end{equation}
}
We discuss the  prior $p\left( \bx, \bz, \bchi \right)$, and  more importantly, approximations to the posterior obtained through Stochastic Variational Inference (SVI) in Section \ref{sec:inference}.
We note that the posterior incorporates all available data, including the governing equations. Furthermore, $\blamc$ enables the identification of collocation points where the postulated constitutive law does not provide adequate closure. Even though it involves not only the material field parameters (i.e. $\bx$) but also those associated with the model state variables (i.e. $\bz,\bchi$), it does {\em not} involve any black-box solver as standard formulations (see \refeqp{eqn:posterior_unnormed}). 
As we demonstrate in the sequel and apart from the advantages above,  this greatly facilitates and expedites the inference process \cite{koutsourelakis2012novel,bruder2018beyond,scholz2025weak}.

\subsection{Probabilistic inference}
\label{sec:inference}
Since the posterior of \refeqp{eq:jposterior} is intractable, we seek approximations in the context of Stochastic Variational Inference (SVI) \cite{blei2017variational}. In particular, we consider a parameterized family of densities $q_{\ve \xi}(\bz, \bx, \bchi, \blamc)$ with tunable parameters $\ve \xi$ and attempt to minimize the Kullback-Leibler (KL) divergence with the exact posterior. This is equivalent to maximizing the Evidence LOwer Bound (ELBO) $\mathcal{L}(\bs{\xi})$ below, where we obmit constant terms \cite{blei2017variational}:
{\small
\begin{equation}
    \begin{array}{ll}
        \log p(\hbRe, \hbRc, \hbu) & = \log \int p \left( \hbu|\bz \right) p\left(\hbRe | \bchi\right) p\left( \hbRc | \bz,\bx, \bchi, \blamc\right) p\left( \bx, \bz, \bchi \right) p(\blamc) ~\diff\bz \diff\bx \diff\bchi \diff \blamc \\
        &\geq \left< \log \frac{p \left( \hbu|\bz \right) p\left(\hbRe | \bchi\right) p\left( \hbRc | \bz,\bx, \bchi, \blamc\right) p\left( \bx, \bz, \bchi \right) p(\blamc)}{q_{\ve \xi}(\bz, \bx, \bchi, \blamc)} \right>_{\qfull} \\ %\quad \textrm{(\person{Jensen}'s inequality)}\\
        & = -\frac{\lame}{2} \sum_{j=1}^{N_e} \left< \left(\re_{j} \right)^2 (\bchi) \right>_{\qfull} \\
        & + \frac{1}{2} \left< \sum_{k=1}^{N_c} \sum_{\ell=1}^{N_\ell} \log \lamc_{k \ell} \right>_{q_{\bs{\xi}}(\bs{\lambda}^{(c)}) } - \sum_{k=1}^{N_c} \sum_{\ell=1}^{N_\ell} \left< \frac{\lamc_{k \ell}}{2}  \left( \rc_{k \ell}  (\bz, \bx, \bchi) \right)^2 \right>_{\qfull} \\
        &- \frac{\tau}{2}  \sum_{i=1}^{N_{\hbu}} \left< (\hu_i -  \u_i(\bz))^2 \right>_{\qfull} \\
        & + \left< \log \frac{p\left(\bz, \bx, \bchi \right) p(\blamc)}{\qfull} \right>_{\qfull} \\
        & = \mathcal{L}(\ve \xi),
    \end{array}
\label{eqn:ELBO}
\end{equation}
}
where $\left< \cdot \right>_{q_{\ve \xi}}$ denotes the expectation with respect to $q_{\ve{\xi}}$. We note the following:
\begin{itemize}
    \item The first term of the ELBO promotes the minimization of the conservation-law residuals.
    \item The second term encourages the minimization of the constitutive-law residuals, with the degree of enforcement governed by the unknown $\lamc_{k \ell}$'s. In particular, the impact of larger residuals in the ELBO, which imply invalidity of the constitutive law, can be damped by smaller precision values $\lamc_{k \ell}$ and vice versa.
    \item The third term minimizes the discrepancy between displacement observations $\hbu$ and the inferred displacements $\bu$. 
    \item The fourth term provides regularization by minimizing the KL-divergence of the approximate posterior with the prior. 
\end{itemize}

Following  \cite{scholz2025weak} and to reduce the update cost, we employ a Monte Carlo approximation of the first ELBO term by subsampling  $n_e \ll N_e$ of the pre-selected weight functions and corresponding residuals as:
\begin{equation}
    \sum_{j=1}^{N_e} \left< \left(\re_{j} \right)^2 (\bchi) \right>_{q_{\ve \xi}(\bz, \bx, \bchi, \blamc)} \approx \frac{{N_e}}{n_e} \sum_{k=1}^{n_e} \left< \left(\re_{j_{k}} \right)^2 (\bchi) \right>_{q_{\ve \xi}(\bz, \bx, \bchi, \blamc)}, \qquad j_{k} \sim \mathrm{Cat}\left({N_e}, \frac{1}{{N_e}} \right). 
    \label{eqn:approx_virt_like}
\end{equation}

The full set of weight functions $\{\bW_j\}_{j=0}^{N_{\bW} }$ to subsample from is created before the simulation. 

\begin{Remarks}
    \item We employ weighting functions $\bW_j$ corresponding to the nodal basis functions of the linear finite element discretization; each $\bW_j$ takes a value of one at node $j$ and decreases linearly to zero across all surrounding elements sharing that node. 
    \item The collocation points for $\bRc$ are selected on a regular grid, details of which are given in the respective numerical experiments of Section \ref{sec:results}. No subsampling,  as in \refeqp{eqn:approx_virt_like} for the residuals of the conservation law, was employed, although this would be possible.
   We note that an adaptive strategy, where the density of collocation points is progressively increased in regions where constitutive-model errors have been detected, could also be readily envisioned, even though it was not used here. 
\end{Remarks}

The approximate posterior $q_{\ve\xi}$ is critical for accuracy and efficiency, requiring a balance between expressiveness and computational simplicity. To motivate our choice, it is important  to note that displacement field $\bu$, the material property  field $\X$, and the stress field $\bsig$ (and thus their respective parameters $\bz$, $\bx$ and $\bchi$) are strongly dependent as they must jointly ensure that the conservation-law residuals $\bRe$ and constitutive-law residuals $\bRc$ in the virtual likelihoods are minimized. We postulate, therefore, an approximate posterior of the form:
\begin{equation}
    \q(\bz, \bx, \bchi, \blamc) = \q(\bx | \bz) \q(\bchi | \bz) \q(\bz) \q(\blamc).
    \label{eqn:qgen}
\end{equation}
To capture the variability in the material fields, we approximate their joint posterior using a combination of simpler, tractable distributions. Specifically, we represent the material, displacement, and stress field parameters as multivariate normal distributions with low-rank covariance structures.
The means of $\q(\bx | \bz)$ and $\q(\bchi | \bz)$ are parameterized using neural networks, which take a shared latent variable $\bz$ as input. 
This amortization 
mapping from a shared latent code to the material and stress field 
parameters encodes the strong physical dependencies between $\bu$, 
$\X$, and $\bsig$ that arise from the governing equations, while 
keeping inference tractable by avoiding the need to optimize a 
separate set of variational parameters for each field independently.
For the hyperparameters $\blamc$, the optimal $\q(\blamc)$ attains the form of  a Gamma distribution, the parameters of which can be updated in closed-form \cite{bishop_variational_2000}. A detailed explanation of these choices, including sampling procedures and distributional assumptions, is provided in Appendix~\ref{app:posterior}.

The gradient of the ELBO with respect to the model parameters is computed using PyTorch's automatic differentiation module \cite{NEURIPS2019_9015}. Parameter updating is performed via ADAM \cite{kingma2014adam} with standard parameters ($\beta_1=0.9, \beta_2 = 0.999$) and learning rate $\rho = 10^{-4}$. The pseudo-code is given in Algorithm \ref{alg:svi}.
We note that multi-sample estimators \cite{mnih_variational_2016}, such as the importance weighted ELBO (IW-ELBO) of \cite{burda_importance_2016}, which provide a tighter bound to the log-evidence with the same computational cost in terms of residual evaluations, offer a promising alternative that was not explored herein. 

Finally, upon convergence and as described in detail in  \ref{app:estimates}, we generate estimates of approximate posterior statistics (i.e., means, variances, and quantiles) using Monte Carlo samples, which are used to assess the reconstruction quality of the material field $\X$ and the quantification of the methods capability to capture constitutive law correctness (or incorrectness). Their values are reported for each of the examples considered in Section \ref{sec:results}.

\begin{algorithm}[t!]
\caption{Stochastic Variational Inference (SVI)}
\label{alg:svi}
\begin{algorithmic}[1]
\State \textbf{Input:} {$\hbu$: displacement observations; $\tau$: observation noise precision $\lame$: conservation precision; $\ve \rho$: learning rate; $n_e$: \#WF/iteration; $\{ \eta^{\x}_i$, $\ve{\eta}^{\bz}_i(\s), \ve{\eta}^{\bchi}_i(\s) \}$: discretization; $\ve f, \ve g$: Boundary conditions; $\s_k$: Collocation point position.}
\State \textbf{Initialize:} Variational parameters $\ve \xi \leftarrow \ve \xi_0$, hyperparameters $a_{k \ell}, b_{k \ell}$, set of weight functions $\bW$, iteration counter $t \leftarrow 0$.
\While{$\mathcal{L}$ not converged}
\State {Subsample $n_e$ weight functions $\{\bW^{(j_h)} : j_h \in [n_e]\}$}
\Statex \hspace{\algorithmicindent} \texttt{\# Sample from variational distributions}
\State {Draw variables $\bz, \bx, \bchi \sim q_{\ve \xi}(\bz, \bx, \bchi)$} \Comment{Eqs. \eqref{eqn:sample_z} -- \eqref{eqn:sample_chi}}
\Statex \hspace{\algorithmicindent} \texttt{\# Compute variational objective}
\State {Evaluate expected precision $\left<\blamc\right>_{\q} = a_{k \ell} / b_{k \ell}$} \Comment{Eq. \eqref{eqn:E_lamc}}
\State {Estimate variational lower bound $\mathcal{L}\left(\ve \xi\right)$} \Comment{Eq. \eqref{eqn:ELBO}}
\Statex \hspace{\algorithmicindent} \texttt{\# Parameter updates}
\State {Compute gradient $\nabla_{\ve \xi} \mathcal{L}\left(\ve \xi\right)$ via automatic differentiation}
\State {Update variational parameters $\ve \xi_{t + 1} \leftarrow \ve \xi_{t} + \ve \rho^{(t)} \odot \nabla_{\ve \xi} \mathcal{L}_{\ve \xi}$}
\State {Update hyperparameters $a_{k \ell}, b_{k \ell}$ for precision estimation} \Comment{Eq. \eqref{eqn:lambdac_closest_form}}
\State {$t \leftarrow t + 1$}
\EndWhile
\State \textbf{Output:} {Optimized variational parameters $\ve \xi$ and learned posterior distributions.}
\end{algorithmic}
\end{algorithm}

\section{Numerical Results} \label{sec:results}
We present two complementary sets of experiments designed to probe 
both the \emph{sensitivity} and the \emph{specificity} of the 
proposed constitutive validity detection capability. The synthetic 
studies (Section~\ref{subsec:brain}) use a controlled two-dimensional 
brain slice geometry in which an anisotropic inclusion is embedded in 
an otherwise linear elastic domain, providing a setting where the 
ground truth validity map is known exactly. Within this setting, we 
progressively challenge the method by degrading data quality through 
increasing noise (Section~\ref{subsec:noise}) and reducing observation 
density down to a single measured displacement component 
(Section~\ref{subsec:data}), testing how robustly the method detects 
constitutive violations as conditions deteriorate. The phantom 
experiment (Section~\ref{subsec:phantom}) addresses the complementary 
question of specificity: given real experimental data from a material 
known to follow linear elastic behavior throughout, does the method 
correctly identify the constitutive model as valid everywhere, without 
generating spurious violations? Together, these experiments 
demonstrate that the method is both sensitive enough to detect 
localized model failures and specific enough to avoid false positives 
under realistic measurement conditions.

\subsection{Synthetic Validation Studies with Brain Slice Geometry} \label{subsec:brain}
\begin{figure}
    \centering
    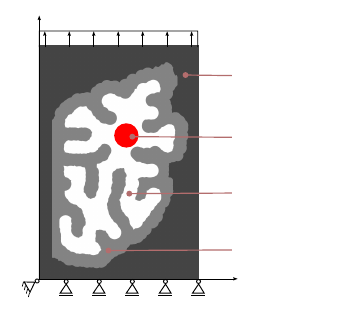
    \caption{Brain slice material property field with boundary conditions. Details about the material properties can be found in Table \ref{tab:material_properties}. The synthetic geometry was inspired by \cite{kamali2023elasticity}.}
    \label{fig:brain_slice}
\end{figure}

We demonstrate the method's core capabilities using a two-dimensional brain slice geometry ($\Omega \subset \mathbb{R}^2$) with heterogeneous material properties designed to simulate realistic tissue contrasts (see Figure \ref{fig:brain_slice}). The geometry is based on \cite{kamali2023elasticity}.

\textbf{Conservation law: } 
Elasticity imaging  of the brain is generally conducted using harmonic excitation at prescribed angular frequency ${\omega = 2 \pi \cdot \qty{50}{\hertz}}$ \cite{hamhaber2010vivo}. All state variables below (i.e. displacements and stresses) can be expressed in the frequency domain \cite{sinkus2000high}. The corresponding conservation law (balance of linear momentum) can be incorporated in the proposed  framework by substituting the weighted residuals of Equation \eqref{eqn:rc} with
\begin{equation}
    \int_{\Omega} \sig_{mn} \W_{m,n}^{(j)} \diff\Omega + \int_{\Omega} \rho \omega^2 \u_{m} \W_{m}^{(j)} \diff\Omega - \int_{\Gamma_N}  f_m \W_m^{(j)} \diff\Gamma_N,
\end{equation}
where $\rho$ is the tissue density (assumed constant at $\rho = 1000$ kg/m$^3$), and $\bs{f}$ represents the harmonic body forces. Henceforth, $\bsig$ and $\bu$ represent the Fourier transforms of the respective fields. Evaluation of spatial integrals is conducted numerically using the FE representation of the fields.

\textbf{Constitutive law:} We distinguish between four different regions, three of which (Background, Gray matter, White matter in Figure \ref{fig:brain_slice}) follow an isotropic, linear-elastic material law
\begin{equation}
    \sig_{ij} = \lambda \delta_{ij} \epsilon_{kk} + 2 \mu \epsilon_{ij},
    \label{eqn:linearConst}
\end{equation}
where $\lambda = E \nu ((1+\nu)(1-2\nu))^{-1}$ and $\mu = E(2(1+\nu))^{-1}$ are the Lam\'{e} parameters. For the fourth region (Inclusion in Fig. \ref{fig:brain_slice}), we employ  an anisotropic material law 
\begin{equation}
    \sig_{ij} = D_{ij} \epsilon_{jk},
    \label{eqn:anisotropic}
\end{equation}
where $\ve D$ is an anisotropic constitutive matrix. The material property values used for the different regions are summarized in Table \ref{tab:material_properties}.While the reference solution is based on the material laws as described, the inverse problem assumes a linear-elastic material behavior throughout the entire domain, as expressed in Equation \eqref{eqn:linearConst}, and aims to determine the distribution of the Young's modulus.

A natural question is whether adopting a more expressive constitutive law with additional unknown parameters would be preferable. We argue that this does not resolve the fundamental issue. As noted in the introduction, increasing model expressiveness may simply obscure invalidity rather than eliminate it: a sufficiently flexible model can approximate observed behavior even without providing any reliable physical interpretation of the inferred parameters. Our primary contribution is demonstrating the ability to detect and localize constitutive model failure for a given assumed law, a capability that remains valuable regardless of how that law is chosen. Practically, simpler constitutive laws also yield more tractable inference: fewer unknown parameters reduce the computational burden and may improve convergence speed, while the validity assessment still signals when the chosen model is insufficient, and refinement is warranted. 

\textbf{Synthetic data generation} 
The synthetic data generation follows the forward model described in Section~\ref{sec:problem}, where displacement observations $\hat{\bu}$ are generated by solving the equilibrium equations with known material properties and subsequently corrupted with additive Gaussian noise according to Equation~\eqref{eqn:obs}. The applied Gaussian noise given by $\tau$ is determined by the signal-to-noise ratio (SNR) in decibels (dB) using formula
\begin{equation}
    \mathrm{SNR}_{\mathrm{dB}} = 10 \log_{10}\left( \left( \frac{\tau}{d_{\bu}} \sum_{i=1}^{d_{\bu}} U_i^2 \right)^{-0.5} \right).
\end{equation}
Detailed model specifications for the forward problem are provided in \ref{app:forward}.

\textbf{Discretization}
For the inverse problem, the domain is discretized on a triangulated $64 \times 64$ grid, with $4096$ collocation points for constitutive residuals $\bRc$ and uniform displacement observations of both field components at $4096$ locations. While material property field $\X$ and each component of the (Fourier-transformed) stress field $\bsig$ are discretized using an element-wise constant, discontinuous elements, the (Fourier-transformed) displacements are discretized using first-order polynomial elements. While Neumann boundary conditions are assumed to be known, the Dirichlet boundary values must be inferred from noisy displacement observations at their prescribed locations.

\textbf{Priors} 
As we have no prior information for the variables $\boldsymbol{\chi}$ and $\by$, we assign an uninformative normal priors $\mathcal{N}(\boldsymbol{0}, 10^{16} \boldsymbol{I})$ . For $\boldsymbol{\lambda}_c$, we employ a gamma prior as specified in Equation~\eqref{eqn:prior_lamc}. For the discretized material properties $\bx$, we employ a prior that penalizes in a learnable manner spatial jumps and promotes piecewise constant solutions. Specifically,  we denote with $\boldsymbol{J}_m = \boldsymbol{B m}$ the vector of differences between neighboring FE elements ("jumps") of dimension $d_{\text{jumps}}$, where $\bs{B}$ is matrix of dimensions $d_{\text{jumps}}  \times d_{\bx}$ that calculate the jumps from the parameters $\bx$. We impose a hierarchical ARD prior on $\boldsymbol{J}_m$ that consists of
\begin{equation}
    \label{eqn:jump_prior}
    \begin{aligned}
        p(\bx | \boldsymbol{\theta}) &= \mathcal{N}(\boldsymbol{J}_m | \boldsymbol{0}, \text{diag}(\boldsymbol{\theta}^{-1})), \quad \mathrm{and} \\
        p(\boldsymbol{\theta}) &= \prod_{j=1}^{d_{\text{jumps}}} \text{Gam}(\theta_j | a_0 = 10^{-6}, b_0 = 10^{-6}).
    \end{aligned}
\end{equation}
The precision hyper-parameters $\boldsymbol{\theta}$ promote sparsity in the number of jumps, i.e., piecewise constant solutions of the material property field $\X$ \cite{bardsley2013gaussian}.  

\textbf{Approximate posterior}
We use the approximate posterior as specified in subsection \ref{sec:inference} and \ref{app:posterior}. We further introduce an additional term associated with the hyperparameter $\boldsymbol{\theta}$ specified in Equations \eqref{eqn:jump_prior} as
\begin{equation}
        \q(\boldsymbol{\theta}) = \prod_{j=1}^{d_{\mathrm{jumps}}} \mathrm{Gam}\left(\theta_j | a_j, b_j \right).
\end{equation}
The neural networks for the approximate posterior in Eq. \eqref{eqn:m_and_sigma} are modeled as a fully-connected multilayer perceptron with three hidden layers, each of them with 2000 neurons, layer normalization, and SiLu activation function.
 
\textbf{Implementation details}
We select a batch size of $n_e = 250$ weight functions per iteration 
for the conservation residuals (Equation~\eqref{eqn:approx_virt_like}). 
Training proceeds in two phases. In the first phase, we fix the 
parameters of $\q(\blamc)$ such that $\blamc = 10^6$ throughout the 
domain and run for $2 \cdot 10^6$ iterations. This initialization 
enforces the constitutive law as a near-hard constraint during the 
early stages of training, allowing the displacement and material 
fields to reach a reasonable initialization before the constitutive 
precision is released. Releasing $\blamc$ prematurely -- before the 
fields are well-initialized -- risks spurious collapse of 
$\blamc$ to near zero in regions where residuals are large for 
reasons unrelated to constitutive model failure (e.g., poor 
initialization of $\bu$ or $\bsig$). In the second phase, 
$\q(\blamc)$ is updated jointly with the remaining variational 
parameters as described in \ref{app:posterior}, and 
training continues until convergence of the ELBO. The learning rate 
is initialized at $3 \cdot 10^{-4}$ and decays to $10^{-4}$ and 
$3 \cdot 10^{-5}$ at iterations $2 \cdot 10^5$ and $10^6$, 
respectively. All computations were performed on an NVIDIA H100 GPU 
with 80~GB of memory.

\begin{table}
\centering
\caption{Material properties for synthetic brain slice geometry.}
\label{tab:material_properties}
\begin{tabular}{lccc}
\hline
Region Name & Material Parameters & Color in Figure & Percentage of Domain \\
\hline
Background & $E = 1.0$ kPa, $\nu = 0.4$ & Dark gray & $43.8$\% \\
Gray matter & $E = 1.4$ kPa, $\nu = 0.4$ & Light gray & $30.5$\% \\
White matter & $E = 1.9$ kPa, $\nu = 0.4$ & White & $24.6$\% \\
Inclusion & $\bs{D} = \begin{bmatrix} 2.0 & -1.0 \\ -1.0 & 1.75 \end{bmatrix}$ kPa & Red & $1.1$\% \\
\hline
\end{tabular}
\end{table}

\subsubsection{Proof of Concept} \label{subsec:poc}

We begin by asking the most fundamental question about the proposed 
framework: when the assumed constitutive law is invalid in a localized 
region, does the method correctly identify where that failure occurs, 
and does it warn that material property estimates in that region 
cannot be trusted diagnostically?

\begin{figure}
    \centering
    \includegraphics{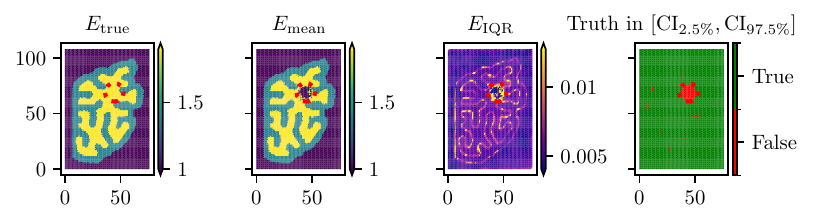}
    \caption{Young's modulus reconstruction for subsection \ref{subsec:poc}: (a) ground truth with red circle where the true material law given by Eq. \eqref{eqn:anisotropic} and thus Young's Modulus is not defined, (b) inferred posterior mean, (c) inferred $95\%$ inter quantile range, (d) coverage map indicating where the $95\%$ credibility intervals cover the ground truth. The dashed circles indicate the region where the assumed constitutive law does not hold.}
    \label{fig:proof_concept_youngs}
\end{figure}

To make the stakes concrete, consider first what any standard inverse 
method -- whether direct, indirect, or learning-based -- would return 
in this setting. Figure~\ref{fig:proof_concept_youngs} shows that the 
inferred posterior mean of the Young's modulus field predicts a 
spuriously soft region in the anisotropic inclusion, where the linear 
elastic constitutive assumption does not hold and where a scalar 
Young's modulus is not physically defined. In the absence of a 
validity indicator, this prediction could be clinically 
misinterpreted: a soft, roughly circular region in brain tissue might 
suggest an arterial structure, an edematous zone, or a low-grade 
lesion. In reality, the apparent softness is an artifact of model 
mismatch -- the inversion is fitting a linear elastic model to tissue 
that does not behave as one, and the result carries no physical 
meaning. This failure mode is not a deficiency of the inversion 
algorithm; it is an unavoidable consequence of optimizing under an 
invalid model assumption, and would arise in any method that does not 
explicitly test constitutive validity.

\begin{figure}
    \centering
    \includegraphics{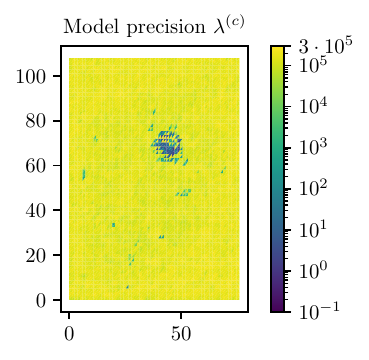}
    \vspace{0.04\textwidth}
    \includegraphics{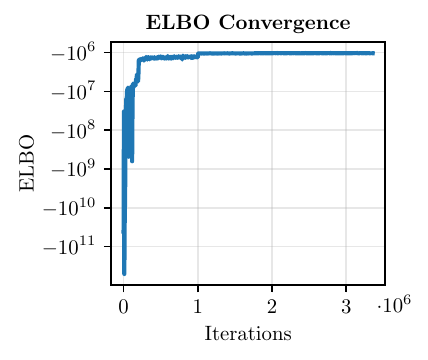}
    \caption{Method validation capabilities for subsection \ref{subsec:poc}: (a) constitutive validity map showing regions of model inadequacy (i.e. $\blamc$) where lower values indicate less trust in the correctness of the assumed material law at a specific location, and (b) ELBO convergence demonstrating optimization stability.}
    \label{fig:proof_concept_validity}
\end{figure}

The proposed framework detects and localizes this failure directly, 
as demonstrated in Figure~\ref{fig:proof_concept_validity}. The 
inferred constitutive precision map $\blamc$ provides a pointwise 
measure of trust in the assumed constitutive law: in the anisotropic 
inclusion, where the linear elastic model cannot produce the stress 
state required by mechanical equilibrium, the precision values 
collapse. In the surrounding domain, where the linear elastic 
assumption holds, the precision values are on average \textbf{five 
orders of magnitude higher} -- a contrast that is robust across the 
noise levels examined in Section~\ref{subsec:noise}. This contrast 
emerges directly from the inference and is not a tuned threshold or 
post-hoc classification: it reflects the degree to which the 
constitutively predicted stress $\tbsig$ can match the 
equilibrium-required stress $\bsig$ at each point.

\begin{figure}
    \centering
    \includegraphics{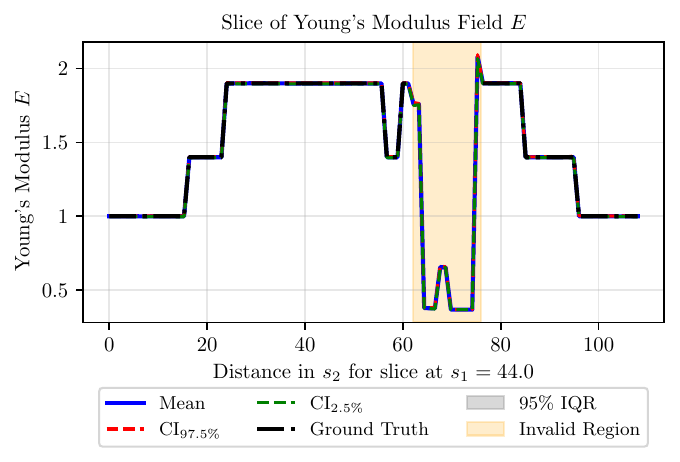}
    \caption{Slice through the Young's Modulus field $E$ at $s_1=44$ mm for the results in subsection \ref{subsec:poc}. We can observe agreement between the true values and the inferred values (lines overlap in the plot). Note that there are no true values in the yellowish region, as the constitutive law does not hold there.}
    \label{fig:E_slice}
\end{figure}

Outside the anisotropic inclusion -- in the $98.9\%$ of the domain 
where the linear elastic assumption holds -- the posterior mean aligns 
closely with the ground truth and the $95\%$ inter-quantile range 
correctly envelopes the true values (Figure~\ref{fig:proof_concept_youngs}). 
Posterior uncertainty is appropriately elevated near material 
boundaries and in regions of higher stiffness, consistent with the 
physical expectation that sharper contrasts are harder to resolve. 
Figure~\ref{fig:E_slice} confirms this through a vertical slice at 
$s_1 = \qty{44}{\milli\meter}$, where the posterior mean and ground 
truth overlap closely throughout the valid domain. The validity map 
thus serves a dual function: it spatially delineates which estimates 
can be acted on clinically and which should be withheld pending model 
refinement or further investigation.

\begin{figure}
    \centering
    \includegraphics{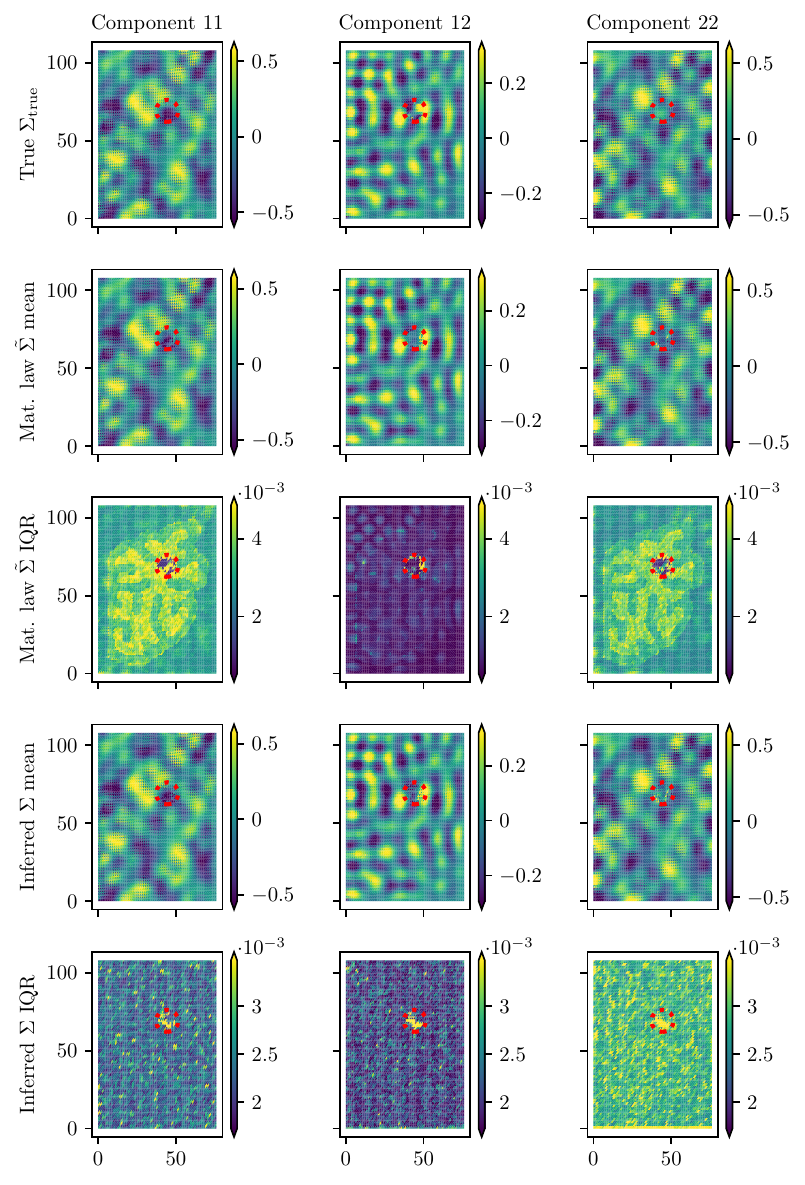}
    \caption{Stress field validation and constitutive model assessment for subsection \ref{subsec:poc}. From left to right, the columns represent the (independent) stress field components $\Sigma_{11}, \Sigma_{12}$ and $\Sigma_{22}$. From top to bottom, the rows depict the ground truth stresses $\Sigma_{\mathrm{true}}$, the constitutive-law-derived mean stress $\tbsig(\bx, \bz)$, its $95\%$ inter quantile range, the   mean of the inferred stress field $\bsig(\bchi) $, and its $95\%$ inter quantile range. The red, dashed circles mark the subdomain where the assumed linear-elastic constitutive law does not hold.}
    \label{fig:stress_validation}
\end{figure}

To understand how this model error detection capability emerges, Figure~\ref{fig:stress_validation} provides a detailed examination of the stress field reconstruction. The key insight lies in recognizing that our framework simultaneously infers two distinct stress representations: the true stress $\bsig(\bchi)$ that satisfies the conservation law (i.e. equations of equilibrium) and the constitutive-law-based stress $\tbsig(\bx, \bz)$.% derived from the assumed material law. 
The inferred true stress components $\bsig(\bchi)$ (rows four and five) closely approximate the ground truth stress $\bsig_{\text{true}}$ (row one) throughout the entire domain, as expected given their role in satisfying the PDE. In contrast, the constitutive-law-derived stress components $\tbsig(\bx, \bz)$ (rows two and three) agree well with both $\bsig(\bchi)$ and $\bsig_{\text{true}}$ only in regions where the linear elastic assumption holds. In the anisotropic inclusion region (highlighted by the red circle), where the constitutive law fails to describe the true material behavior, $\tbsig$ exhibits significant deviations from both $\bsig$ and $\bsig_{\text{true}}$ across all stress components. This discrepancy between the mechanically required stress and the constitutively predicted stress directly manifests as the model error captured by the low precision values in Figure~\ref{fig:proof_concept_validity} (left). As such, the framework reveals where the assumed constitutive law cannot produce the stress state required by mechanical equilibrium, thereby identifying regions of model invalidity.
Convergence is validated in Figure~\ref{fig:proof_concept_validity} (right), where the ELBO reaches a stable value after approximately $3.5 \cdot 10^6$ iterations.

\subsubsection{Robustness to Measurement Noise} \label{subsec:noise}
\begin{figure}[!h]
    \centering
    \includegraphics{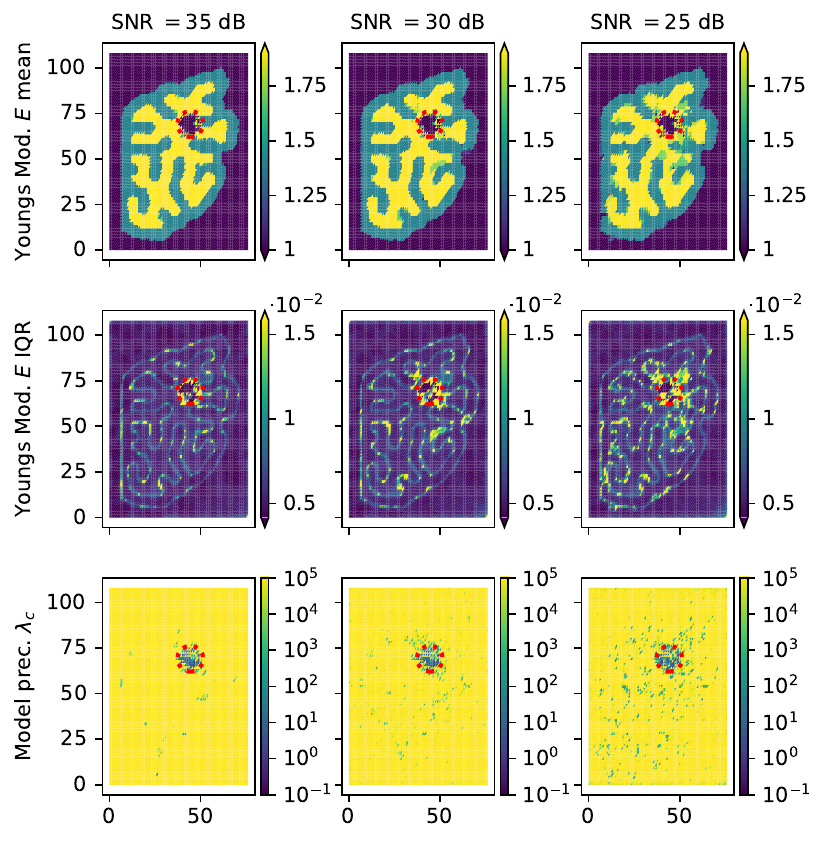}
    \caption{Impact of measurement noise on reconstruction and uncertainty quantification from subsection \ref{subsec:noise}: posterior means (first row) and inter quantile range (second row) of Young's modulus field $\X$ and inferred model precision $\blamc$ (third row) at SNR levels of 35, 30, and 25 dB (columns one to three, respectively).}
    \label{fig:noise_uncertainty}
\end{figure}

Figure~\ref{fig:noise_uncertainty} demonstrates that the method appropriately adapts its uncertainty quantification to degraded data quality. The posterior means of the material property field $\X(\bs{s})$ remain visually consistent with the ground truth across all noise levels, with uncertainty systematically increasing with noise level, most visibly at material boundaries. This boundary-concentrated uncertainty is a natural consequence of the smoothness prior, which penalizes jumps in the material field and thereby keeps within-region uncertainty relatively low even at 25 dB.
This stability is partly attributable to the conservation law 
residuals, which provide dense physical constraints throughout the 
domain independently of the displacement observations: even as 
measurement noise increases, the requirement that the inferred stress 
field satisfies mechanical equilibrium continues to anchor the 
reconstruction, partially compensating for the loss of information 
in the noisy observations.

The constitutive violation detection similarly adapts to noise, as shown in the third row of Figure~\ref{fig:noise_uncertainty}. At 35 dB, the method sharply distinguishes valid from invalid regions. At 25 dB, the inclusion remains clearly identifiable as the dominant cluster of constitutive inadequacy, though isolated single-element artifacts appear near material boundaries. These artifacts are consistent with the behavior of the smoothness prior on the material field $\X$ (i.e., the Young's Modulus field $E$): near boundaries, neither neighboring material region provides a perfect local fit, and the resulting discrepancy is interpreted as constitutive error rather than material heterogeneity. More broadly, inferred trust in the constitutive law decreases across the domain as noise increases, reflecting the fundamental limitation that noisier observations provide less definitive evidence for model adequacy.
Importantly, however, the inclusion remains the dominant and most 
spatially coherent cluster of low precision across all noise levels 
tested, confirming that the five-orders-of-magnitude contrast 
reported in Section~\ref{subsec:poc} is not an artifact of low-noise 
conditions but reflects a robust signal that persists under realistic 
clinical noise levels.

\subsubsection{Impact of Data Sampling Density and Partial Displacement Field} \label{subsec:data}

We examine robustness under two realistic data acquisition constraints. First, we reduce the number of observed displacement points from four data points per displacement unknown ($100\%$ sampling) to one ($25\%$ sampling), retaining both displacement components. Second, we test a single-component scenario motivated by medical elastography, where certain imaging modalities can only measure displacements in one direction \cite{luo2009effects}. Here, only $\u_2(\bs{s})$ is observed while $\u_1(\bs{s})$ remains entirely unmeasured, corresponding to $12.5\%$ data availability relative to the baseline. In all cases, we maintain a noise level of $30$ dB and an uninformative prior on $\bu$, so that reconstruction relies solely on the governing equations and observed data.

\begin{figure}
    \centering
    \includegraphics{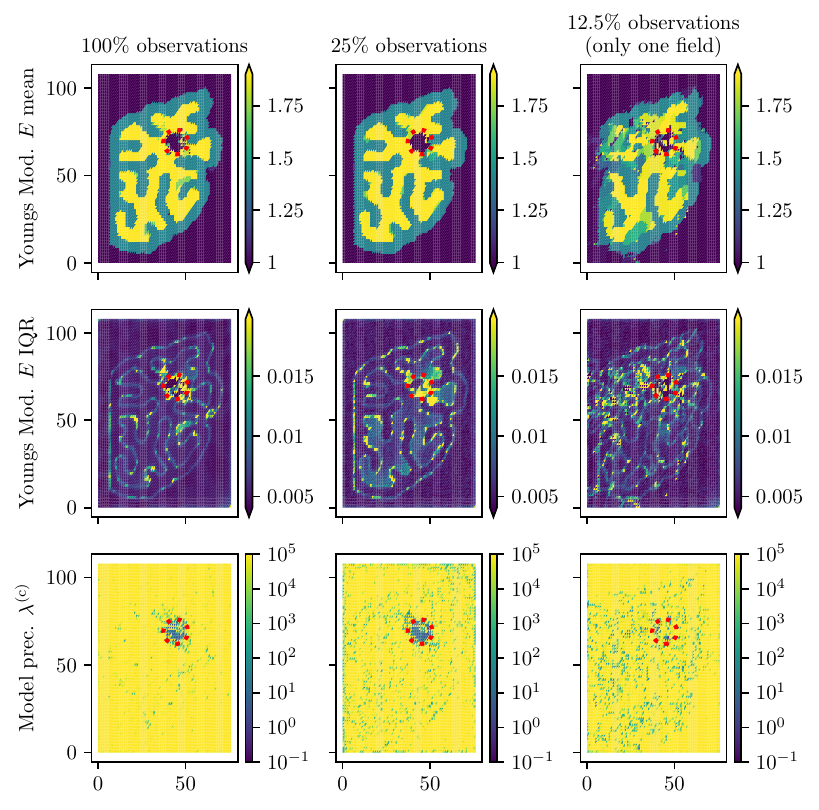}
    \caption{Impact of number of available data points on reconstruction and uncertainty quantification at 30 dB noise level from subsection \ref{subsec:data}: posterior means (first row) and inter quantile range (second row) of Young's modulus field $\X$ and inferred model precision $\blamc$ (third row) with $100\%$ ($4:1$ data points versus unknowns in $\bz$),  $25\%$ (1:1), and $12.5\%$ (0.5:1, where field $U_1$ is unobserved) in columns one to three, respectively.}
    \label{fig:sampling_reconstruction}
\end{figure}

Figure~\ref{fig:sampling_reconstruction} presents the results. Comparing the $100\%$ and $25\%$ sampling scenarios reveals only minor differences in the posterior mean and interquartile range of $E$, with the inclusion remaining clearly identifiable as a cluster of constitutively invalid material. The method thus exhibits robustness under a four-fold reduction in observation density. With only a single displacement component available (right column), the main features of the material field are still recovered through the physics-based constraints, though with degraded inference quality. More critically, constitutive violation detection loses sensitivity: the inclusion is no longer distinguishable from the surrounding domain. As the method invokes parsimony through the prior (see Section~\ref{sec:main}), it prefers solutions where the prescribed material law is valid at as many points as possible, and therefore infers a field $\u_1(\bs{s})$ consistent with the constitutive model throughout the domain. This inferred field necessarily differs from the true but unmeasured ground truth, in which the constitutive violation is present, thereby masking the anomaly within the inclusion.

These findings suggest that, at least in the two-dimensional setting 
considered here, constitutive violation detection benefits 
substantially from multi-component displacement measurements: with 
only a single component available, the method cannot distinguish 
between a genuine constitutive violation and a displacement field that 
is simply consistent with the assumed model in the measured direction. 
Whether this limitation persists in three-dimensional settings -- 
where additional displacement components may provide the missing 
constraints -- is an open question that warrants further investigation. 
In the meantime, experimental studies in which constitutive model validation is a primary objective should, where possible, prioritize acquisition protocols that provide displacement measurements in multiple directions.

\subsection{Experimental Validation with Phantom Data} \label{subsec:phantom}
Having validated the method on synthetic cases and demonstrated its ability to detect constitutive model error, we now turn to a complementary concern: the risk of false positives, i.e., detecting model error where none exists. Such fraudulent detections could arise, for instance, under elevated noise levels, due to discretization error, or when the noise deviates from the assumed Gaussian model. To assess this risk, we consider an experimental phantom with known, valid linear elastic behavior, and ask whether the method correctly identifies the material behavior as consistent, while simultaneously recovering accurate material parameters.

\subsubsection{Experiment Setup}
We use experimental data collected for and provided by the École Centrale de Lyon \cite{seppecher2023reconstructing} on a commercially available elastography phantom of type "CIRS Model 049" from Computerized Imaging Reference Systems in Norfolk (VA, USA).

\begin{figure}[!t]
    \centering
    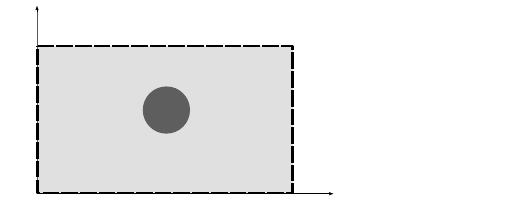
    \caption{CIRS phantom 049 experimental setup: sketch of the domain with stiff inclusion within a constant background, where boundary conditions are unknown. Reported are the values at the Region of Interest. All measures in $\qty{}{\milli\meter}$.}
    \label{fig:phantom_setup}
\end{figure}

\textbf{Geometry}. The phantom  measures \qty{40}{\milli\meter} $\times$ \qty{60}{\milli\meter} and contains a single spherical inclusion of diameter \qty{10}{\milli\meter} (see Figure~\ref{fig:phantom_setup}). 

\textbf{Conservation law:} The governing equation is the balance of linear momentum as in Equation~\eqref{eqn:rc}. No boundary conditions, neither Dirichlet nor Neumann, are available. Rather than resorting to penalty terms and smoothing \cite{seppecher2023reconstructing}, our framework handles this naturally by {\em inferring} boundary displacement values on $\partial\Omega$ during inference while  the weight functions employed are set to zero on $\partial\Omega$ \cite{scholz2025weak}.

\textbf{Constitutive Law:} Following \cite{seppecher2023reconstructing}, we invoke the plane stress assumption, reducing the constitutive behavior to
\begin{equation}
    \begin{bmatrix} \sigma_{11} \\ \sigma_{22} \\ \sigma_{12} \end{bmatrix}
    =
    \frac{E}{1-\nu^2}
    \begin{bmatrix} 1 & \nu & 0 \\ \nu & 1 & 0 \\ 0 & 0 & \frac{1-\nu}{2} \end{bmatrix}
    \begin{bmatrix} \varepsilon_{11} \\ \varepsilon_{22} \\ 2\varepsilon_{12} \end{bmatrix}.
    \label{eqn:plane_stress}
\end{equation}
The inclusion has a Young's modulus of \qty{47}{\kilo\pascal} against a background of \qty{26.5}{\kilo\pascal}, yielding a true stiffness ratio of $\eta_\true = 1.77$. Both regions follow linear elastic behavior with Poisson's ratio $\nu = 0.495$ throughout domain $\Omega$.

\begin{figure}
    \centering
    \includegraphics[width=0.99\textwidth]{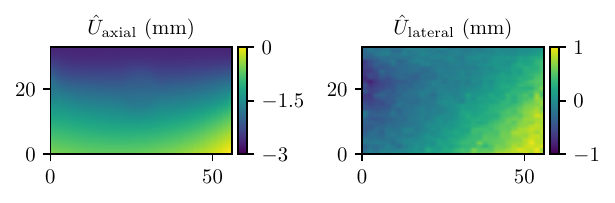}
    \caption{Observed displacements $\hat{\bu}$ of the CIRS Model 049 phantom for lateral and axial displacements. Note that the laterial measured displacements are significantly more noisy than the axial ones.}
    \label{fig:phantom_displacements}
\end{figure}

\textbf{Discretization and Measurements:} Displacement measurements are available over a \qty{56}{\milli\meter} $\times$ \qty{33}{\milli\meter}   region (Figure~\ref{fig:phantom_displacements}), from which we extract a square Region of Interest of \qty{22}{\milli\meter} $\times$ \qty{22}{\milli\meter} with $64 \times 64$ observations in both axial and lateral directions. The Region of Interest is discretized on a triangulated $32 \times 32$ grid with the same kind of finite element shape function as in subsection \ref{subsec:brain}. As is inherent to ultrasound-based displacement tracking, the measured field $\hat{\bu}$ exhibits direction-dependent noise, with substantially higher uncertainty in the lateral direction than the axial with $\tau_{\text{axial}}^{-0.5} = \qty{1e-3}{\milli\meter}$ and $\tau_{\text{lateral}}^{-0.5} = \qty{1e-1}{\milli\meter}$. These are accounted for via two separate likelihood terms.

\textbf{Priors:}
In contrast to the smoothing prior on the material field $\X$ employed in subsection~\ref{subsec:brain}, we adopt a vague normal prior on $\log(\bx)$ as $\mathcal{N}(\textbf{0}, 2\mathbf{I})$. Since the experiment is normalized such that the reference stiffness is unity (see below "Implementation Details"), this prior reflects the mild assumption that true stiffness values are unlikely to deviate from the reference by more than an order of magnitude, i.e., roughly $[e^{-2}, e^{2}]$ on the reference scale. This choice deliberately places minimal informative structure on the material field, demonstrating that the method can recover accurate reconstructions even in the absence of strong prior knowledge.

\textbf{Implementation Details:} Since absolute stiffness cannot be recovered without Neumann boundary conditions, we fix the stiffness to $1$ (unitless) in a \qty{5}{\milli\meter}-wide strip at the bottom of the Region of Interest (see Figure \ref{fig:phantom_case1}) as an internal reference.

\subsubsection{Phantom Results}
\begin{figure}[!t]
    \centering
    \includegraphics[width=0.99\textwidth]{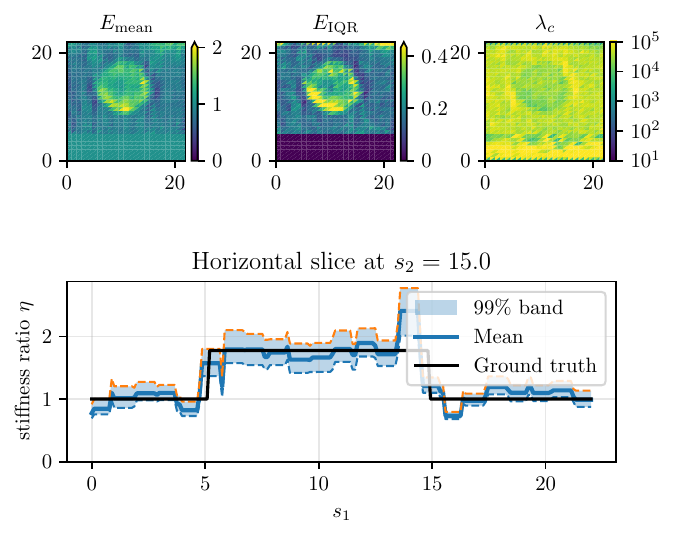}
    \caption{Results for the CIRS Model 049 reconstruction in subsection \ref{subsec:phantom}. Top row from left to right showing the (1) approximate posterior mean for the relative Young's Modulus $E_\mathrm{mean}$, (2) its interquartile range $E_\mathrm{IQR}$, and (3) the inferred constitutive law precision $\blamc$ confirms high model adequacy throughout the domain. Bottom plot shows the Young's Modulus stiffness ratio $\eta$ along a slice at $y=15$ mm with ground truth (black), mean (blue), and quantile band (light blue) covering $99\%$ probability mass. }
    \label{fig:phantom_case1}
\end{figure}

The primary question this experiment is designed to answer is whether 
the method generates false positive constitutive violations on data 
from a material known to follow linear elastic behavior. 
Figure~\ref{fig:phantom_case1} (top right) shows the inferred 
constitutive precision map $\blamc$: values of $\lambda^{(c)}_k \sim 
\mathcal{O}(10^5)$ are obtained throughout the entire domain, 
including near the inclusion boundary and in the noisier lateral 
direction, where spurious violations might be expected. No region 
exhibits the precision collapse observed in the synthetic anisotropic 
inclusion (Section~\ref{subsec:poc}). The method therefore correctly 
identifies the constitutive model as globally adequate,  producing no 
false positives despite the elevated lateral noise, unknown 
boundary conditions, and real measurement artifacts inherent to 
ultrasound-based displacement tracking.

Having established specificity, we turn to quantitative accuracy. 
The posterior mean of the Young's modulus field 
(Figure~\ref{fig:phantom_case1}, top left) clearly resolves the stiff 
inclusion against the uniform background, with appropriately elevated 
uncertainty near the inclusion boundary in the interquantile range 
plot (top middle). The bottom strip of zero uncertainty reflects the 
fixed reference region. Figure~\ref{fig:phantom_case1} (bottom) shows 
a horizontal slice at $s_2 = \qty{15}{\milli\meter}$ through the 
stiffness ratio field $\eta = E / E_{\text{ref}}$, where $E_{\text{ref}}$ 
is chosen such that $\eta \approx 1$ in the background. Since absolute 
stiffness is not recoverable without Neumann boundary conditions, this 
relative scaling leaves all stiffness ratios intact. The true value 
$\eta_{\mathrm{true}} = 1.77$ falls within the $99\%$ credibility 
interval across the inclusion, confirming that the method recovers 
both the location and relative stiffness of the inclusion with good 
quantitative accuracy, even under the challenging conditions of this 
experiment.

\section{Discussion and Outlook}
We have presented a probabilistic framework for elastography that 
treats constitutive model validity as an explicit inference target 
rather than an implicit assumption. By introducing the stress field 
as an independent latent variable and enforcing the conservation law 
and constitutive law through separate virtual likelihoods with 
distinct precision hyperparameters, the framework enables a pointwise 
comparison between the stress required by mechanical equilibrium and 
the stress predicted by the assumed constitutive model. The resulting 
spatially resolved precision field $\blamc$ provides an 
uncertainty-aware map of where the constitutive assumption is 
supported by the data and where it is not -- a capability that, to 
our knowledge, has not previously been demonstrated in a fully 
probabilistic elastography framework. This represents a fundamental 
shift from traditional elastography paradigms, where model adequacy 
is assumed rather than assessed, and addresses a risk that affects 
direct and indirect methods, Bayesian and deterministic formulations, 
and learning-based approaches equally: that an invalid constitutive 
assumption will produce estimates that appear plausible but carry no 
reliable physical interpretation.

The experimental results establish two key properties of the 
framework. First, \emph{sensitivity}: in the synthetic brain slice 
geometry, the inferred precision field $\blamc$ correctly identifies 
the anisotropic inclusion -- where the assumed linear elastic model 
is invalid -- with a contrast of five orders of magnitude relative 
to the surrounding valid domain. This signal is robust to noise 
levels spanning the range from controlled laboratory conditions to 
realistic clinical scenarios (SNR 35 to 25 dB), and persists under 
a four-fold reduction in observation density. Second, 
\emph{specificity}: in the phantom experiment using real 
ultrasound-based displacement measurements from a material known to 
follow linear elastic behavior, the framework produces no false 
positive constitutive violations -- the precision field remains 
uniformly high throughout the domain despite elevated lateral noise, 
unknown boundary conditions, and real measurement artifacts. 
Simultaneously, the method recovers the relative stiffness contrast 
between inclusion and background with the true value falling within 
the $99\%$ credibility interval, confirming that constitutive 
validity assessment does not come at the cost of reconstruction 
accuracy. The forward-model-free inference via stochastic variational 
inference makes the approach computationally tractable without 
repeated forward solves, and the framework extends naturally to 
different governing equations, as demonstrated through both static 
and time-harmonic elastography formulations.

Several limitations warrant acknowledgment. The most fundamental 
is that $\blamc$ measures the degree to which the assumed 
constitutive law can produce stresses consistent with mechanical 
equilibrium, but cannot on its own attribute a detected mismatch 
to any specific cause. Constitutive model failure, discretization 
error, and unmodeled boundary condition effects can all produce 
reduced precision values, and the framework does not disambiguate 
between them. The phantom experiment provides some evidence that 
discretization error at the resolution used does not generate 
spurious violations, but this cannot be guaranteed in general. 
A practical implication is that the precision field should be 
interpreted in conjunction with other diagnostic information -- 
for example, checking whether detected violations are spatially 
correlated with mesh boundaries or regions of uncertain boundary 
conditions -- rather than as a standalone classifier of 
constitutive model failure.

A second limitation concerns the two-phase training schedule, 
in which the constitutive precision is held fixed during an initial 
warm-up phase before being released for joint inference. While this 
prevents premature collapse of $\blamc$ during early training, 
the length of the warm-up phase and the initial value of 
$\blamc$ are problem-dependent choices that currently require 
manual selection. An adaptive or self-tuning strategy for 
this initialization would improve the usability of the framework 
in new application settings. Finally, the absolute scale of 
$\blamc$ depends on the normalization of the constitutive 
residuals and cannot be directly compared across problems with 
different domains, discretizations, or physical scales without 
appropriate rescaling,  a consideration that should be taken 
into account when applying the framework to new settings.

Several directions present themselves for future development. The 
most pressing is developing principled methods to attribute detected 
precision reductions to specific causes , i.e.  constitutive model 
failure, discretization error, or boundary condition uncertainty, 
potentially by combining the proposed framework with approximation 
error methods \cite{kaipio2006statistical} that can pre-characterize 
the statistical signature of discretization and truncation errors. 
Extension to time-dependent and transient elastography formulations 
would broaden applicability to dynamic tissue characterization, and 
is a natural next step given that the virtual observable framework 
accommodates general PDE residuals. Given evidence that viscoelastic 
models can help distinguish malignant from benign tumors 
\cite{qiu2008ultrasonic}, incorporating viscoelastic constitutive 
laws (and assessing their validity spatially) represents a 
high-impact clinical development direction. Adaptive collocation 
point placement, concentrating collocation points in regions where 
precision values are low, could improve both computational efficiency 
and detection sensitivity in an iterative refinement strategy. The 
framework's structure is not specific to elastography: any 
PDE-constrained inverse problem in which conservation laws can be 
trusted but closure relations cannot -- including subsurface 
geomechanics, cardiovascular mechanics, and structural health 
monitoring -- presents the same opportunity for constitutive 
validity assessment, and extending the framework to these domains 
is a natural direction. Finally, the discretization-free formulation 
outlined in \ref{app:latent} offers a path to eliminating 
mesh-dependency, which would remove one potential confound from the 
precision field and strengthen its interpretability as a pure 
indicator of constitutive model adequacy.

\section*{Competing Interests}
The authors declare that they have no known competing financial interests or personal relationships that could have influenced the work reported in this paper.

\section*{Acknowledgements}
The authors would like to thank Laurent Seppecher from
École Centrale de Lyon for providing the measurement data from the elastography phantom.

\section*{Funding}
The support of the Deutsche Forschungsgemeinschaft (DFG) through Project Number 499746055 is gratefully acknowledged. 

\section*{Author Contributions}
\textbf{Vincent C. Scholz}: Writing – review \& editing, Writing – original draft, Visualization, Validation, Software, Methodology, Investigation, Formal analysis, Data curation, Conceptualization. \textbf{P.S. Koutsourelakis}: Writing – review \& editing, Writing – original draft, Supervision, Resources, Project administration, Methodology, Funding acquisition, Formal analysis, Conceptualization.

\section*{Data Availability}
The code used in this study will be made publicly available on GitHub upon publication of this paper at: \\
\url{https://github.com/pkmtum/Model-Error-Quantification-for-WNVI}.

% ================== BIB ==================

\printbibliography

@article{banerjee_large_2013,
        title = {Large scale parameter estimation problems in frequency-domain elastodynamics using an error in constitutive equation functional},
        volume = {253},
        issn = {0045-7825}, 
        url = {https://www.sciencedirect.com/science/article/pii/S0045782512002770},
        doi = {10.1016/j.cma.2012.08.023},
        urldate = {2026-04-09},
        journal = {Computer Methods in Applied Mechanics and Engineering},
        author = {Banerjee, Biswanath and Walsh, Timothy F. and Aquino, Wilkins and Bonnet, Marc},
        month = jan,
        year = {2013},        pages = {60--72},
}

@article{bonnet2005inverse,
  title     = {Inverse problems in elasticity},
  author    = {Bonnet, Marc and Constantinescu, Andrei},
  journal   = {Inverse Problems},
  volume    = {21},
  number    = {2},
  pages     = {R1},
  year      = {2005},
  publisher = {IOP Publishing}
}

@article{patel_circumventing_2019,
        title = {Circumventing the solution of inverse problems in mechanics through deep learning: {Application} to elasticity imaging},
        volume = {353},
        issn = {0045-7825},
        shorttitle = {Circumventing the solution of inverse problems in mechanics through deep learning},
        url = {https://www.sciencedirect.com/science/article/pii/S0045782519302579},
        doi = {10.1016/j.cma.2019.04.045},
        urldate = {2026-03-29},
        journal = {Computer Methods in Applied Mechanics and Engineering},
        author = {Patel, Dhruv and Tibrewala, Raghav and Vega, Adriana and Dong, Li and Hugenberg, Nicholas and Oberai, Assad A.},
        month = aug,
        year = {2019},
        pages = {448--466},
}

@article{babaniyi_direct_2017,
        series = {Special {Issue} on {Biological} {Systems} {Dedicated} to {William} {S}. {Klug}},
        title = {Direct error in constitutive equation formulation for plane stress inverse elasticity problem},
        volume = {314},
        issn = {0045-7825},
        url = {https://www.sciencedirect.com/science/article/pii/S0045782516313834},
        doi = {10.1016/j.cma.2016.10.026},
        urldate = {2026-03-29},
        journal = {Computer Methods in Applied Mechanics and Engineering},
        author = {Babaniyi, Olalekan A. and Oberai, Assad A. and Barbone, Paul E.},
        month = feb,
        year = {2017},
        pages = {3--18},
        }

@article{feissel_modified_2007,
        title = {Modified constitutive relation error identification strategy for transient dynamics with corrupted data: {The} elastic case},
        volume = {196},
        issn = {0045-7825},
        shorttitle = {Modified constitutive relation error identification strategy for transient dynamics with corrupted data},
        url = {https://www.sciencedirect.com/science/article/pii/S0045782506003434},
        doi = {10.1016/j.cma.2006.10.005},
        number = {13},
        urldate = {2026-03-05},
        journal = {Computer Methods in Applied Mechanics and Engineering},
        author = {Feissel, P. and Allix, O.},
        month = mar,
        year = {2007},
        pages = {1968--1983},
}

@article{dasgupta_conditional_2025,
        title = {Conditional score-based diffusion models for solving inverse elasticity problems},
        volume = {433},
        issn = {0045-7825},
        url = {https://www.sciencedirect.com/science/article/pii/S0045782524006807},
        doi = {10.1016/j.cma.2024.117425},
        pages = {117425},
        journaltitle = {Computer Methods in Applied Mechanics and Engineering},
        shortjournal = {Computer Methods in Applied Mechanics and Engineering},
        author = {Dasgupta, Agnimitra and Ramaswamy, Harisankar and Murgoitio-Esandi, Javier and Foo, Ken Y. and Li, Runze and Zhou, Qifa and Kennedy, Brendan F. and Oberai, Assad A.},
        urldate = {2025-07-24},
        date = {2025-01-01},
}

@article{zhang_solution_2012,
        title = {Solution of the time-harmonic viscoelastic inverse problem with interior data in two dimensions},
        volume = {92},
        issn = {1097-0207},
        url = {https://onlinelibrary.wiley.com/doi/abs/10.1002/nme.4372},
        doi = {10.1002/nme.4372},
        language = {en},
        number = {13},
        urldate = {2025-06-05},
        journal = {International Journal for Numerical Methods in Engineering},
        author = {Zhang, Yixiao and Oberai, Assad A and Barbone, Paul E and Harari, Isaac},
        year = {2012},
        note = {\_eprint: https://onlinelibrary.wiley.com/doi/pdf/10.1002/nme.4372},
        pages = {1100--1116},
}

@article{hoerig_data-driven_2019,
        title = {Data-{Driven} {Elasticity} {Imaging} {Using} {Cartesian} {Neural} {Network} {Constitutive} {Models} and the {Autoprogressive} {Method}},
        volume = {38},
        issn = {1558-254X},
        url = {https://ieeexplore.ieee.org/abstract/document/8522049},
        doi = {10.1109/TMI.2018.2879495},
        number = {5},
        urldate = {2025-06-04},
        journal = {IEEE Transactions on Medical Imaging},
        author = {Hoerig, Cameron and Ghaboussi, Jamshid and Insana, Michael F.},
        month = may,
        year = {2019},
        pages = {1150--1160},
}

@article{tripura_wavelet_2023,
        title = {A wavelet neural operator based elastography for localization and quantification of tumors},
        volume = {232},
        issn = {0169-2607},
        url = {https://www.sciencedirect.com/science/article/pii/S0169260723001037},
        doi = {10.1016/j.cmpb.2023.107436},
        urldate = {2025-06-04}, 
        journal = {Computer Methods and Programs in Biomedicine},
        author = {Tripura, Tapas and Awasthi, Abhilash and Roy, Sitikantha and Chakraborty, Souvik},
        month = apr,
        year = {2023},
        pages = {107436},
}

@article{ormachea_elastography_2020,
        title = {Elastography imaging: the 30 year perspective},
        volume = {65},
        issn = {0031-9155},
        shorttitle = {Elastography imaging},
        url = {https://dx.doi.org/10.1088/1361-6560/abca00},
        doi = {10.1088/1361-6560/abca00},
        language = {en},
        number = {24},
        urldate = {2025-06-04},
        journal = {Physics in Medicine \& Biology},
        author = {Ormachea, J and Parker, K J},
        month = dec,
        year = {2020},
        note = {Publisher: IOP Publishing},
        pages = {24TR06},
}

@article{fovargue_stiffness_2018,
        title = {Stiffness reconstruction methods for {MR} elastography},
        volume = {31},
        copyright = {© 2018 The Authors. NMR in Biomedicine published by John Wiley \& Sons Ltd.},
        issn = {1099-1492},
        url = {https://onlinelibrary.wiley.com/doi/abs/10.1002/nbm.3935},
        doi = {10.1002/nbm.3935},
        language = {en},
        number = {10},
        urldate = {2025-06-04},
        journal = {NMR in Biomedicine}, 
        author = {Fovargue, Daniel and Nordsletten, David and Sinkus, Ralph},
        year = {2018},
        note = {\_eprint: https://onlinelibrary.wiley.com/doi/pdf/10.1002/nbm.3935},
        pages = {e3935},
}

@article{sack_magnetic_2022,
        title = {Magnetic resonance elastography from fundamental soft-tissue mechanics to diagnostic imaging},
        volume = {5},
        issn = {2522-5820}, 
        url = {https://www.nature.com/articles/s42254-022-00543-2},
        doi = {10.1038/s42254-022-00543-2}, 
        language = {en},
        number = {1},
        urldate = {2025-06-04},
        journal = {Nature Reviews Physics},
        author = {Sack, Ingolf},
        month = nov,
        year = {2022},
        pages = {25--42},
}

@book{finlayson_method_1972,
        address = {New York},
        title = {The method of weighted residuals and variational principles, with application in fluid mechanics, heat and mass transfer, {Volume} 87},
        isbn = {978-0-12-257050-6},
        language = {English},
        publisher = {Academic Press},
        editor = {Finlayson, BA},
        month = feb,
        year = {1972},
}

@article{biehler_towards_2015,
	title = {Towards efficient uncertainty quantification in complex and large-scale biomechanical problems based on a {Bayesian} multi-fidelity scheme},
	volume = {14},
	issn = {1617-7940},
	url = {https://doi.org/10.1007/s10237-014-0618-0},
	doi = {10.1007/s10237-014-0618-0},
	language = {en},
	number = {3},
	urldate = {2025-05-07},
	journal = {Biomechanics and Modeling in Mechanobiology},
	author = {Biehler, Jonas and Gee, Michael W. and Wall, Wolfgang A.},
	month = jun,
	year = {2015},
	pages = {489--513},
	}

@article{goenezen_linear_2012,
        title = {Linear and nonlinear elastic modulus imaging: an application to breast cancer diagnosis},
        volume = {31},
        issn = {1558-254X},
        shorttitle = {Linear and nonlinear elastic modulus imaging},
        doi = {10.1109/TMI.2012.2201497},
        language = {eng},
        number = {8},
        journal = {IEEE transactions on medical imaging},
        author = {Goenezen, Sevan and Dord, Jean-Francois and Sink, Zac and Barbone, Paul E. and Jiang, Jingfeng and Hall, Timothy J. and Oberai, Assad A.},
        month = aug,
        year = {2012},
        pmid = {22665504},
        pmcid = {PMC3698046},
        pages = {1628--1637},
}

@article{oberai2003solution,
  title={Solution of inverse problems in elasticity imaging using the adjoint method},
  author={Oberai, Assad A and Gokhale, Nachiket H and Feij{\'o}o, Gonzalo R},
  journal={Inverse problems},
  volume={19},
  number={2},
  pages={297--313},
  year={2003}
}

@article{barbone_estimating_2011,
        title = {Estimating uncertainty in inverse elasticity with application to quantitative elastography},
        volume = {130},
        issn = {0001-4966},
        url = {https://doi.org/10.1121/1.3654643},
        doi = {10.1121/1.3654643},
        number = {4\_Supplement},
        urldate = {2023-06-29},
        journal = {The Journal of the Acoustical Society of America},
        author = {Barbone, Paul E. and Chue, Bryan and Oberai, Assad A.},
        month = oct,
        year = {2011}, 
        pages = {2405},
}

@inproceedings{carvalho_handling_2009,
        title = {Handling sparsity via the horseshoe},
        booktitle = {International {Conference} on {Artificial} {Intelligence} and {Statistics}},
        author = {Carvalho, Carlos M and Polson, Nicholas G and Scott, James G},
        year = {2009},
        pages = {73--80},
}

@article{ishwaran_spike_2005,
        title = {Spike and slab variable selection: {Frequentist} and {Bayesian} strategies},
        volume = {33},
        issn = {0090-5364, 2168-8966},
        shorttitle = {Spike and slab variable selection},
        url = {http://projecteuclid.org/euclid.aos/1117114335},
        doi = {10.1214/009053604000001147},
        number = {2},
        urldate = {2016-06-01},
        journal = {The Annals of Statistics},
        author = {Ishwaran, Hemant and Rao, J. Sunil},
        month = apr,
        year = {2005},
        mrnumber = {MR2163158},
        zmnumber = {1068.62079},
        pages = {730--773}, 
}

@article{burda_importance_2016,
        title = {Importance {Weighted} {Autoencoders}},
        url = {http://arxiv.org/abs/1509.00519},
        language = {en},
        urldate = {2022-05-11},
        journal = {arXiv:1509.00519 [cs, stat]},
        author = {Burda, Yuri and Grosse, Roger and Salakhutdinov, Ruslan},
        month = nov,
        year = {2016},
        note = {arXiv: 1509.00519},
}

@misc{mnih_variational_2016,
        title = {Variational inference for {Monte} {Carlo} objectives},
        url = {http://arxiv.org/abs/1602.06725},
        doi = {10.48550/arXiv.1602.06725},
        urldate = {2025-03-26},
        publisher = {arXiv},
        author = {Mnih, Andriy and Rezende, Danilo J.}, 
        month = jun,
        year = {2016}, 
        note = {arXiv:1602.06725 [cs]},
        }

@article{fuhg_review_2024,
        title = {A {Review} on {Data}-{Driven} {Constitutive} {Laws} for {Solids}},
        issn = {1886-1784},
        url = {https://doi.org/10.1007/s11831-024-10196-2},
        doi = {10.1007/s11831-024-10196-2},
        language = {en},
        urldate = {2025-03-26}, 
        journal = {Archives of Computational Methods in Engineering},
        author = {Fuhg, Jan N. and Anantha Padmanabha, Govinda and Bouklas, Nikolaos and Bahmani, Bahador and Sun, WaiChing and Vlassis, Nikolaos N. and Flaschel, Moritz and Carrara, Pietro and De Lorenzis, Laura},
        month = nov,
        year = {2024},
}

@Article{Bayarri2009,
	author     = {M. J. Bayarri and J. O. Berger and F. Liu},
	journal    = {Bayesian Analysis},
	title      = {Modularization in Bayesian analysis, with emphasis on analysis of computer models},
	year       = {2009},
	month      = mar,
	number     = {1},
	volume     = {4},
	doi        = {10.1214/09-ba404},
	publisher  = {Institute of Mathematical Statistics},
}

@Article{Plumlee2017,
	author     = {Matthew Plumlee},
	journal    = {Journal of the American Statistical Association},
	title      = {Bayesian Calibration of Inexact Computer Models},
	year       = {2017},
	month      = jun,
	number     = {519},
	pages      = {1274--1285},
	volume     = {112},
	doi        = {10.1080/01621459.2016.1211016},
	publisher  = {Informa {UK} Limited},
}

@Article{Leoni2024,
	author    = {Nicolas Leoni and Olivier Le Ma{\^{\i}}tre and Maria-Giovanna Rodio and Pietro Marco Congedo},
	journal   = {International Journal for Uncertainty Quantification},
	title     = {Bayesian Calibration With Adpative Model Discrepancy},
	year      = {2024},
	number    = {1},
	pages     = {19--41},
	volume    = {14},
	doi       = {10.1615/int.j.uncertaintyquantification.2023046331},
	publisher = {Begell House},
}

@Article{Brynjarsdottir2014,
	author    = {Jenn{\'{y}} Brynjarsd{\'{o}}ttir and Anthony O'Hagan},
	journal   = {Inverse Problems},
	title     = {Learning about physical parameters: the importance of model discrepancy},
	year      = {2014},
	month     = oct,
	number    = {11},
	pages     = {114007},
	volume    = {30},
	doi       = {10.1088/0266-5611/30/11/114007},
	publisher = {{IOP} Publishing},
}

@incollection{mackay_bayesian_1996,
        title = {Bayesian methods for backpropagation networks},
        booktitle = {Models of {Neural} {Networks} {III}}, 
        publisher = {Springer},
        author = {MacKay, DJC},
        editor = {Domany, E and van Hemmen, JL and Schulten, K},
        year = {1996},
        pages = {211--254},
}

@inproceedings{bishop_variational_2000,
        title = {Variational {Relevance} {Vector} {Machines}},
        booktitle = {{UAI}},
        author = {Bishop, Christopher M. and Tipping, Michael E.},
        year = {2000},
        pages = {46--53},
}

@article{doyley2012model,
  title={Model-based elastography: a survey of approaches to the inverse elasticity problem},
  author={Doyley, Marvin M},
  journal={Physics in Medicine \& Biology},
  volume={57},
  number={3},
  pages={R35},
  year={2012},
  publisher={IOP Publishing}
}

@article{ophir1991elastography,
  title={Elastography: a quantitative method for imaging the elasticity of biological tissues},
  author={Ophir, Jonathan and Cespedes, Ignacio and Ponnekanti, Hari and Yazdi, Youseph and Li, Xin},
  journal={Ultrasonic imaging},
  volume={13},
  number={2},
  pages={111--134},
  year={1991},
  publisher={SAGE Publications Sage CA: Los Angeles, CA}
}

@article{muthupillai1995magnetic,
  title={Magnetic resonance elastography by direct visualization of propagating acoustic strain waves},
  author={Muthupillai, R and Lomas, DJ and Rossman, PJ and Greenleaf, James F and Manduca, Armando and Ehman, Richard Lorne},
  journal={science},
  volume={269},
  number={5232},
  pages={1854--1857},
  year={1995},
  publisher={American Association for the Advancement of Science}
}

@article{khalil2005tissue,
  title={Tissue elasticity estimation with optical coherence elastography: toward mechanical characterization of in vivo soft tissue},
  author={Khalil, Ahmad S and Chan, Raymond C and Chau, Alexandra H and Bouma, Brett E and Mofrad, Mohammad R Kaazempur},
  journal={Annals of biomedical engineering},
  volume={33},
  pages={1631--1639},
  year={2005},
  publisher={Springer}
}

@article{raissi2019physics,
  title={Physics-informed neural networks: A deep learning framework for solving forward and inverse problems involving nonlinear partial differential equations},
  author={Raissi, Maziar and Perdikaris, Paris and Karniadakis, George E},
  journal={Journal of Computational physics},
  volume={378},
  pages={686--707},
  year={2019},
  publisher={Elsevier}
}

@article{haghighat2021physics,
  title={A physics-informed deep learning framework for inversion and surrogate modeling in solid mechanics},
  author={Haghighat, Ehsan and Raissi, Maziar and Moure, Adrian and Gomez, Hector and Juanes, Ruben},
  journal={Computer Methods in Applied Mechanics and Engineering},
  volume={379},
  pages={113741},
  year={2021},
  publisher={Elsevier}
}

@article{van2025enforcing,
  title={Enforcing physics onto PINNs for more accurate inhomogeneous material identification},
  author={van der Heijden, B and Li, X and Lubineau, Gilles and Florentin, E},
  journal={Computer Methods in Applied Mechanics and Engineering},
  volume={441},
  pages={117993},
  year={2025},
  publisher={Elsevier}
}

@article{xu2023transfer,
  title={Transfer learning based physics-informed neural networks for solving inverse problems in engineering structures under different loading scenarios},
  author={Xu, Chen and Cao, Ba Trung and Yuan, Yong and Meschke, G{\"u}nther},
  journal={Computer Methods in Applied Mechanics and Engineering},
  volume={405},
  pages={115852},
  year={2023},
  publisher={Elsevier}
}

@article{zang2020weak,
  title={Weak adversarial networks for high-dimensional partial differential equations},
  author={Zang, Yaohua and Bao, Gang and Ye, Xiaojing and Zhou, Haomin},
  journal={Journal of Computational Physics},
  volume={411},
  pages={109409},
  year={2020},
  publisher={Elsevier}
}

@article{kamali2023elasticity,
  title={Elasticity imaging using physics-informed neural networks: Spatial discovery of elastic modulus and Poisson's ratio},
  author={Kamali, Ali and Sarabian, Mohammad and Laksari, Kaveh},
  journal={Acta biomaterialia},
  volume={155},
  pages={400--409},
  year={2023},
  publisher={Elsevier}
}

@inproceedings{mcgarry2011comparison,
  title={Comparison of iterative and direct inversion MR elastography algorithms},
  author={McGarry, MDJ and van Houten, EE W and Pattison, AJ and Weaver, JB and Paulsen, KD},
  booktitle={Mechanics of Biological Systems and Materials, Volume 2: Proceedings of the 2011 Annual Conference on Experimental and Applied Mechanics},
  pages={49--56},
  year={2011},
  organization={Springer}
}

@article{sumi1995estimation,
  title={Estimation of shear modulus distribution in soft tissue from strain distribution},
  author={Sumi, Chikayoshi and Suzuki, Akifumi and Nakayama, Kiyoshi},
  journal={IEEE Transactions on Biomedical Engineering},
  volume={42},
  number={2},
  pages={193--202},
  year={1995},
  publisher={IEEE}
}

@article{manduca2001magnetic,
  title={Magnetic resonance elastography: non-invasive mapping of tissue elasticity},
  author={Manduca, Armando and Oliphant, Travis E and Dresner, M Alex and Mahowald, JL and Kruse, Scott A and Amromin, E and Felmlee, Joel P and Greenleaf, James F and Ehman, Richard L},
  journal={Medical image analysis},
  volume={5},
  number={4},
  pages={237--254},
  year={2001},
  publisher={Elsevier}
}

@article{manduca2003spatio,
  title={Spatio-temporal directional filtering for improved inversion of MR elastography images},
  author={Manduca, Armando and Lake, David S and Kruse, Scott A and Ehman, Richard L},
  journal={Medical image analysis},
  volume={7},
  number={4},
  pages={465--473},
  year={2003},
  publisher={Elsevier}
}

@article{green2015bayesian,
  title={Bayesian computation: a summary of the current state, and samples backwards and forwards},
  author={Green, Peter J and {\L}atuszy{\'n}ski, Krzysztof and Pereyra, Marcelo and Robert, Christian P},
  journal={Statistics and Computing},
  volume={25},
  pages={835--862},
  year={2015},
  publisher={Springer}
}

@inproceedings{risholm2011probabilistic,
  title={Probabilistic elastography: estimating lung elasticity},
  author={Risholm, Petter and Ross, James and Washko, George R and Wells, William M},
  booktitle={Information Processing in Medical Imaging: 22nd International Conference, IPMI 2011, Kloster Irsee, Germany, July 3-8, 2011. Proceedings 22},
  pages={699--710},
  year={2011},
  organization={Springer}
}

@article{blei2017variational,
  title={Variational inference: A review for statisticians},
  author={Blei, David M and Kucukelbir, Alp and McAuliffe, Jon D},
  journal={Journal of the American statistical Association},
  volume={112},
  number={518},
  pages={859--877},
  year={2017},
  publisher={Taylor \& Francis}
}

@article{franck_multimodal_2017,
        title = {Multimodal, high-dimensional, model-based, {Bayesian} inverse problems with applications in biomechanics},
        volume = {329},
        issn = {0021-9991},
        url = {http://www.sciencedirect.com/science/article/pii/S002199911630537X},
        doi = {10.1016/j.jcp.2016.10.039},
        urldate = {2017-03-07},
        journal = {Journal of Computational Physics}, 
        author = {Franck, I. M. and Koutsourelakis, P. S.},
        month = jan,
        year = {2017},
        pages = {91--125},
}

@article{kingma2013auto,
  title={Auto-encoding variational bayes},
  author={Kingma, Diederik P and Welling, Max},
  journal={arXiv preprint arXiv:1312.6114},
  year={2013}
}

@article{hoffman2013stochastic,
  title={Stochastic variational inference},
  author={Hoffman, Matthew D and Blei, David M and Wang, Chong and Paisley, John},
  journal={Journal of machine learning research},
  year={2013}
}

@incollection{holzapfel2017similarities,
  title={Similarities between soft biological tissues and rubberlike materials},
  author={Holzapfel, Gerhard A},
  booktitle={Constitutive models for rubber IV},
  pages={607--617},
  year={2017},
  publisher={Routledge}
}

@book{kaipio2006statistical,
  title={Statistical and computational inverse problems},
  author={Kaipio, Jari and Somersalo, Erkki},
  volume={160},
  year={2006},
  publisher={Springer Science \& Business Media}
}

@article{kennedy2001bayesian,
  title={Bayesian calibration of computer models},
  author={Kennedy, Marc C and O'Hagan, Anthony},
  journal={Journal of the Royal Statistical Society: Series B (Statistical Methodology)},
  volume={63},
  number={3},
  pages={425--464},
  year={2001},
  publisher={Wiley Online Library}
}

@article{pernot2017critical,
  title={A critical review of statistical calibration/prediction models handling data inconsistency and model inadequacy},
  author={Pernot, Pascal and Cailliez, Fabien},
  journal={AIChE Journal},
  volume={63},
  number={10},
  pages={4642--4665},
  year={2017},
  publisher={Wiley Online Library}
}

@article{sargsyan2015statistical,
  title={On the statistical calibration of physical models},
  author={Sargsyan, Khachik and Najm, Habib N and Ghanem, Roger},
  journal={International Journal of Chemical Kinetics},
  volume={47},
  number={4},
  pages={246--276},
  year={2015},
  publisher={Wiley Online Library}
}

@article{kass1995bayes,
  title={Bayes factors},
  author={Kass, Robert E and Raftery, Adrian E},
  journal={Journal of the american statistical association},
  volume={90},
  number={430},
  pages={773--795},
  year={1995},
  publisher={Taylor \& Francis}
}

@article{arcones2024embedded,
  title={Embedded Model Bias Quantification with Measurement Noise for Bayesian Model Calibration},
  author={Arcones, Daniel Andr{\'e}s and Weiser, Martin and Koutsourelakis, Phaedon-Stelios and Unger, J{\"o}rg F},
  journal={arXiv preprint arXiv:2410.12037},
  year={2024}
}

@inproceedings{newman2024improving,
  title={Improving image quality in a new method of data-driven elastography},
  author={Newman, Will and Ghaboussi, Jamshid and Insana, Michael},
  booktitle={Medical Imaging 2024: Ultrasonic Imaging and Tomography},
  volume={12932},
  pages={51--56},
  year={2024},
  organization={SPIE}
}

@article{hoerig2021machine,
  title={Machine Learning in Model-free Mechanical Property Imaging: Novel Integration of Physics With the Constrained Optimization Process},
  author={Hoerig, Cameron and Ghaboussi, Jamshid and Wang, Yiliang and Insana, Michael F},
  journal={Frontiers in Physics},
  volume={9},
  pages={600718},
  year={2021},
  publisher={Frontiers Media SA}
}

@article{joshi2022bayesian,
  title={Bayesian-EUCLID: Discovering hyperelastic material laws with uncertainties},
  author={Joshi, Akshay and Thakolkaran, Prakash and Zheng, Yiwen and Escande, Maxime and Flaschel, Moritz and De Lorenzis, Laura and Kumar, Siddhant},
  journal={Computer Methods in Applied Mechanics and Engineering},
  volume={398},
  pages={115225},
  year={2022},
  publisher={Elsevier}
}

@article{flaschel_automated_2023-1,
        title = {Automated discovery of generalized standard material models with {EUCLID}},
        volume = {405},
        issn = {0045-7825},
        url = {https://www.sciencedirect.com/science/article/pii/S0045782522008234},
        doi = {10.1016/j.cma.2022.115867},
        pages = {115867},
        journaltitle = {Computer Methods in Applied Mechanics and Engineering},
        shortjournal = {Computer Methods in Applied Mechanics and Engineering},
        author = {Flaschel, Moritz and Kumar, Siddhant and De Lorenzis, Laura},
        urldate = {2025-11-17}, 
        date = {2023-02-15},
}

@article{thakolkaran2022nn,
  title={NN-EUCLID: Deep-learning hyperelasticity without stress data},
  author={Thakolkaran, Prakash and Joshi, Akshay and Zheng, Yiwen and Flaschel, Moritz and De Lorenzis, Laura and Kumar, Siddhant},
  journal={Journal of the Mechanics and Physics of Solids},
  volume={169},
  pages={105076},
  year={2022},
  publisher={Elsevier}
}

@article{man2011neural,
  title={Neural network constitutive modelling for non-linear characterization of anisotropic materials},
  author={Man, Hou and Furukawa, Tomonari},
  journal={International journal for numerical methods in engineering},
  volume={85},
  number={8},
  pages={939--957},
  year={2011},
  publisher={Wiley Online Library}
}

@article{wang2021inference,
  title={Inference of deformation mechanisms and constitutive response of soft material surrogates of biological tissue by full-field characterization and data-driven variational system identification},
  author={Wang, Zhenlin and Estrada, Jonathan B and Arruda, Ellen M and Garikipati, Krishna},
  journal={Journal of the Mechanics and Physics of Solids},
  volume={153},
  pages={104474},
  year={2021},
  publisher={Elsevier}
}

@article{koutsourelakis2012novel,
  title={A novel Bayesian strategy for the identification of spatially varying material properties and model validation: an application to static elastography},
  author={Koutsourelakis, Phaedon-Stelios},
  journal={International Journal for Numerical Methods in Engineering},
  volume={91},
  number={3},
  pages={249--268},
  year={2012},
  publisher={Wiley Online Library}
}

@article{bruder2018beyond,
  title={Beyond black-boxes in Bayesian inverse problems and model validation: applications in solid mechanics of elastography},
  author={Bruder, Lukas and Koutsourelakis, Phaedon-Stelios},
  journal={International Journal for Uncertainty Quantification},
  volume={8},
  number={5},
  year={2018},
  publisher={Begel House Inc.}
}

@article{scholz2025weak,
  title={Weak neural variational inference for solving Bayesian inverse problems without forward models: applications in elastography},
  author={Scholz, Vincent C and Zang, Yaohua and Koutsourelakis, Phaedon-Stelios},
  journal={Computer Methods in Applied Mechanics and Engineering},
  volume={433},
  pages={117493},
  year={2025},
  publisher={Elsevier}
}

@article{kaltenbach2020incorporating,
  title={Incorporating physical constraints in a deep probabilistic machine learning framework for coarse-graining dynamical systems},
  author={Kaltenbach, Sebastian and Koutsourelakis, Phaedon-Stelios},
  journal={Journal of Computational Physics},
  volume={419},
  pages={109673},
  year={2020},
  publisher={Elsevier}
}

@article{vadeboncoeur2023fully,
  title={Fully probabilistic deep models for forward and inverse problems in parametric PDEs},
  author={Vadeboncoeur, Arnaud and Akyildiz, {\"O}mer Deniz and Kazlauskaite, Ieva and Girolami, Mark and Cirak, Fehmi},
  journal={Journal of Computational Physics},
  volume={491},
  pages={112369},
  year={2023},
  publisher={Elsevier}
}

@article{bardsley2013gaussian,
  title={GAUSSIAN MARKOV RANDOM FIELD PRIORS FOR INVERSE PROBLEMS.},
  author={Bardsley, Johnathan M and Kaipio, Jari},
  journal={Inverse Problems \& Imaging},
  volume={7},
  number={2},
  year={2013},
  publisher={Citeseer}
}

@article{kingma2014adam,
  title={Adam: A method for stochastic optimization},
  author={Kingma, Diederik P and Ba, Jimmy},
  journal={arXiv preprint arXiv:1412.6980},
  year={2014}
}

@incollection{NEURIPS2019_9015,
title = {PyTorch: An Imperative Style, High-Performance Deep Learning Library},
author = {Paszke, Adam and Gross, Sam and Massa, Francisco and Lerer, Adam and Bradbury, James and Chanan, Gregory and Killeen, Trevor and Lin, Zeming and Gimelshein, Natalia and Antiga, Luca and Desmaison, Alban and Kopf, Andreas and Yang, Edward and DeVito, Zachary and Raison, Martin and Tejani, Alykhan and Chilamkurthy, Sasank and Steiner, Benoit and Fang, Lu and Bai, Junjie and Chintala, Soumith},
booktitle = {Advances in Neural Information Processing Systems 32},
pages = {8024--8035},
year = {2019},
publisher = {Curran Associates, Inc.},
url = {http://papers.neurips.cc/paper/9015-pytorch-an-imperative-style-high-performance-deep-learning-library.pdf}
}

@article{sinkus2000high,
  title={High-resolution tensor MR elastography for breast tumourdetection},
  author={Sinkus, R and Lorenzen, J and Schrader, D and Lorenzen, M and Dargatz, M and Holz, D},
  journal={Physics in Medicine \& Biology},
  volume={45},
  number={6},
  pages={1649},
  year={2000},
  publisher={IOP Publishing}
}

@article{hamhaber2010vivo,
  title={In vivo magnetic resonance elastography of human brain at 7 T and 1.5 T},
  author={Hamhaber, Uwe and Klatt, Dieter and Papazoglou, Sebastian and Hollmann, Maurice and Stadler, J{\"o}rg and Sack, Ingolf and Bernarding, Johannes and Braun, J{\"u}rgen},
  journal={Journal of Magnetic Resonance Imaging},
  volume={32},
  number={3},
  pages={577--583},
  year={2010},
  publisher={Wiley Online Library}
}

@article{luo2009effects,
  title={Effects of various parameters on lateral displacement estimation in ultrasound elastography},
  author={Luo, Jianwen and Konofagou, Elisa E},
  journal={Ultrasound in medicine \& biology},
  volume={35},
  number={8},
  pages={1352--1366},
  year={2009},
  publisher={Elsevier}
}

@article{seppecher2023reconstructing,
  title={Reconstructing the spatial distribution of the relative shear modulus in quasi-static ultrasound elastography: plane stress analysis},
  author={Seppecher, Laurent and Bretin, Elie and Millien, Pierre and Petrusca, Lorena and Brusseau, Elisabeth},
  journal={Ultrasound in Medicine \& Biology},
  volume={49},
  number={3},
  pages={710--722},
  year={2023},
  publisher={Elsevier}
}

@article{qiu2008ultrasonic,
  title={Ultrasonic viscoelasticity imaging of nonpalpable breast tumors: preliminary results},
  author={Qiu, Yupeng and Sridhar, Mallika and Tsou, Jean K and Lindfors, Karen K and Insana, Michael F},
  journal={Academic radiology},
  volume={15},
  number={12},
  pages={1526--1533},
  year={2008},
  publisher={Elsevier}
}

@article{agrawal_probabilistic_2024,
        title = {A probabilistic, data-driven closure model for {RANS} simulations with aleatoric, model uncertainty},
        volume = {508},
        issn = {0021-9991},
        url = {https://www.sciencedirect.com/science/article/pii/S0021999124002316},
        doi = {10.1016/j.jcp.2024.112982},
        urldate = {2024-11-07},
        journal = {Journal of Computational Physics},
        author = {Agrawal, Atul and Koutsourelakis, Phaedon-Stelios},
        month = jul,
        year = {2024},
        pages = {112982},
}

% ================== APPENDIX ==================

\appendix

\section{Discretization details and alternative formulation} \label{app:latent}
\subsection{Discretization}
We use a finite-dimensional representation (similar to the finite element method) of the material field $\X$ and the independent state variables, i.e., the displacement field $\bu$,
\begin{equation}
    \X(\bx, \s) = \sum_{i=1}^{d_{\x}} \x_i \eta^{\x}_i(\s),  \quad \mathrm{and} \quad \bu(\bz, \s) = \sum_{i=1}^{d_{\bz}} \z_i \ve{\eta}^{\bz}_i(\s) \label{eqn:fields} % \NN_{\xi}\left( \z, \s \right), 
\end{equation}
as well as for the dependent state variables, i.e., the stress field $\bsig$:
\be
\bsig(\ve \chi, \s) = \sum_{i=1}^{d_{\ve \chi}} \chi_i \ve{\eta}^{\bchi}_i(\s).
\label{eqn:sfield} 
\ee
In the expressions above, $\s$ denotes the spatial coordinates and $\eta^{\x}_i$, $\ve{\eta}^{\bz}_i(\s)$ and $\ve{\eta}^{\bchi}_i(\s)$ are given basis functions dependent on space. The coefficients , $\bz = \{ \z_i \}_{i=1}^{d_{\z}}$, $\bx = \{ \x_i \}_{i=1}^{d_{\x}}$ and  $\ve \chi = \{ \chi_i \}_{i=1}^{d_{\ve \chi}}$ represent the discretized versions of the aforementioned fields. Alternative representations involving e.g. neural networks could also be employed as e.g. in PINNs \cite{raissi2019physics} and the associated parameters can be updated by back-propagating the gradients through the respective ELBO terms.

\section{Approximate posterior} \label{app:posterior}
We assume an approximate posterior as in Equation \eqref{eqn:qgen}. Furthermore, we assume that 
\begin{equation}
    \q (\bz) = \mathcal{N}\left( \bz | \ve \mu_{\bz}, \ve S_{\bz} \right).
\end{equation}
Analogously to \cite{scholz2025weak}, we link the mean $\ve{\mu}_{\bx;\ve \xi_{\bx}}$ and $\ve{\mu}_{\ve \chi;\ve \xi_{\bchi}}$ to $\bz$ via deep neural networks (details are problem specific and are discussed in section \ref{sec:results}) with parameters $\ve \xi_{\bx}, \ve \xi_{\bchi}$
\begin{equation}
    \q(\bx | \bz) = \mathcal{N}\left( \bx | ~\ve{\mu}_{\bx;\ve \xi_{\bx}}\left( \bz \right), ~\ve{S}_{\bx} \right) 
    \quad \mathrm{and} \quad 
    \q(\ve \chi | \bz) = \mathcal{N}\left( \ve \chi | ~\ve{\mu}_{\ve \chi;\ve \xi_{\bchi}}\left( \bz \right), ~\ve{S}_{\bchi} \right),
    \label{eqn:m_and_sigma}
\end{equation}
where the conditional covariance $\ve{S}_{\bx}$ and $\ve{S}_{\bchi}$ are assumed to be independent of $\bz$ and of the form
\begin{equation}
    \ve{S}_{\bx}=\ve L_{\bx} \ve L_{\bx}^T + \mathrm{diag}\left( \ve \sigma_{\bx}^2 \right) 
    \quad \mathrm{and} \quad 
    \ve{S}_{\bchi}=\ve L_{\bchi} \ve L_{\bchi}^T + \mathrm{diag}\left( \ve \sigma_{\bchi}^2 \right),
\end{equation}
where $\ve{L}_{\bx}$ (or similarly $ \ve L_{\bchi}$) is a matrix of dimension $d_{\bx} \times d_{\tbx}$ (or $d_{\ve \chi} \times d_{\tilde{\ve \chi}}$) that captures the principal directions along which (conditional) variance is larger. The term  $\ve{\sigma}_{\bx}^2$ (or $\ve{\sigma}_{\bchi}^2$) is a vector of dimension $d_{\bx}$ (or $d_{\bchi}$) that captures the residual (conditional) variance along the $\bx-$ dimensions (or $\ve \chi-$ dimensions).
In contrast to a full covariance matrix, the form adopted for $\ve{S}_{\bx}$ (or $\ve{S}_{\bchi}$) ensures linear scaling of the unknown parameters with $d_{\bx}$ (or $d_{\bchi}$), which for most problems can be high.

Furthermore, it can be readily shown that the optimal approximate posterior for each $\lamc_{k \ell}$ is also a Gamma distribution with parameters $a_{k \ell},b_{k \ell}$, which can be updated in closed form, i.e.
\begin{equation}
    \q(\blamc) = \prod_{k=1}^{N_c} \prod_{\ell=1}^{N_\ell} \mathrm{Gam}(\lamc_{k \ell}|~a_{k \ell},b_{k \ell}),
    \label{eq:qlambda}
\end{equation}
where
\begin{equation}
    a_{k \ell} = a_0 + \frac{1}{2}, \quad \mathrm{and}  \quad b_{k \ell} = b_0 +  \frac{1}{2}\left<  \left( \rc_{k \ell}  (\bz, \bx, \ve \chi) \right)^2  \right>_{\q},
    \label{eqn:lambdac_closest_form}
\end{equation}
and following this the expectation 
\begin{equation}
    \left<  \lamc_{k \ell}  \right>_{\q} = \frac{a_{k \ell}}{b_{k \ell}}.
    \label{eqn:E_lamc}
\end{equation}

 Given this, the  vector of parameters $\ve \xi$ that has to be optimized by maximizing the ELBO consists of:
\begin{equation}
    \ve \xi = \{ \ve \mu_{\bz}, \ve S_{\bz}, \ve \xi_{\bx}, \ve L_{\bx}, \ve \sigma_{\bx}^2, \ve \xi_{\bchi}, \ve L_{\bchi}, \ve \sigma_{\bchi}^2 \}. % \ve a, \ve b \}. % ,  \ve \xi_u
\end{equation}

The ELBO $\mathcal{L}(\ve \xi)$ is maximized using Stochastic Gradient Ascent (SGA), which relies on Monte Carlo estimates of the gradient with respect to $\bs{\xi}$. 
To obtain the latter, we employ the reparameterization trick \cite{kingma2013auto}, which, given the structure of $\q$, entails expressing the variables involved as follows:
\begin{equation}
    \bz = \ve \mu_{\bz} + \ve{S}_{\bz} \ve \varepsilon_1 \quad \mathrm{with} \quad \ve \varepsilon_1 \sim \mathcal{N}\left(\ve 0, \ve I_{d_{\ve z}}\right).
    \label{eqn:sample_z}
\end{equation}
Given this $\bz$, we can proceed to express $\bx$ and $\ve \chi$ as:
\begin{align}
    \bx =\ve{\mu}_{\bx;\ve \xi_{\bx}}\left( \bz \right) + \ve{L}_{\bx} \ve \varepsilon_2+ \ve \sigma_{\bx}  \odot \ve \varepsilon_3, \quad  & \ve \varepsilon_2 \sim \mathcal{N}\left(\ve 0, \ve I_{d_{\tilde{\ve x}}}\right) \mathrm{ \ and \ } \ve \varepsilon_3 \sim \mathcal{N}\left(\ve 0, \ve I_{d_{\ve x}}\right) \quad \mathrm{and} \label{eqn:sample_x}\\
    \ve \chi =\ve{\mu}_{\ve \chi;\ve \xi_{\bchi}}\left( \bz \right) + \ve{L}_{\bchi} \ve \varepsilon_4+ \ve \sigma_{\bchi}  \odot \ve \varepsilon_5, \quad  & \ve \varepsilon_4 \sim \mathcal{N}\left(\ve 0, \ve I_{d_{\tilde{\ve \chi}}}\right) \mathrm{ \ and \ } \ve \varepsilon_5 \sim \mathcal{N}\left(\ve 0, \ve I_{d_{\ve \chi}}\right), \label{eqn:sample_chi}
\end{align}
respectively.

We use $L$  Monte Carlo samples to approximate the expectations with respect to $q_{\bs{\xi}}$ for the weighted residuals $\re$  and collocation-type residuals $\rc$ in \refeqp{eqn:ELBO}. 

\section{Evaluation Metrics} \label{app:estimates}
\subsection{Posterior Estimates}

After convergence, we sample \( B = 1000 \) tuples \( \{ \bz, \bx, \bchi \} \) from the approximate posterior \( \q \); the mean \( \left<\blamc\right>_{\q} = a_{k \ell} / b_{k \ell} \) can be directly obtained. We can then evaluate the fields \( \{ \bu , \X, \bsig \} \) from the respective parameters using Eq. \eqref{eqn:fields} and \eqref{eqn:sfield}. Finally, we estimate the posterior mean and variance via a location-point-wise Monte Carlo estimate of a quantity, e.g., material field $\X$,
\begin{align}
    \label{eqn:mean}
    \mathbb{E}[\X(\ve s) | \hbu, \hbRc, \hbRe] &\approx \frac{1}{B} \sum_{b=1}^B \X(\bx_b,\ve s) \quad \mathrm{and} \\
    \mathrm{Var}[ \X(\ve s) | \hbu, \hbRc, \hbRe ] &\approx \frac{1}{B } \sum_{b=1}^B \left( \X(\bx_b,\ve s) - \bar{\X}(\ve s) \right)^2. \label{eqn:variance}
\end{align}
or estimate the credibility intervals (e.g. $2.5 \%$ and $97.5 \%$) at each location as
\begin{align}
   Q_{0.025} (\X(\ve s)) %= lb_\X(\ve s) 
   &= \mathrm{quantile}(\X_b(\ve s), 0.025) \quad \mathrm{and} \label{eqn:lower_bound}\\
    Q_{0.975} (\X(\ve s)) %=ub_\X(\ve s) 
    &= \mathrm{quantile}(\X_b(\ve s), 0.975).  \label{eqn:upper_bound}
\end{align}
Using the quantiles, we can establish the $95\%$ inter-quantile range (IQR) as
\begin{equation}
    \mathrm{IQR} = Q_{0.975} (\X(\ve s)) - Q_{0.025} (\X(\ve s)).
\end{equation}
which provides an estimate of the posterior uncertainty. 

\subsection{Constitutive Law Correctness}
To evaluate and visualize the spatial distribution of model precision, we consider
the model precision $\lamc_{k\ell}$ of each independent stress component $\ell$
at a given collocation point $\mathbf{s}_k$. Since the components are treated as
independent, the aggregate model variance at $\mathbf{s}_k$ is obtained by summing
the per-component variances,
\begin{equation}
    \frac{1}{\lamc_k} = \sum_{\ell=1}^{N_\ell} \frac{1}{\lamc_{k\ell}},
\end{equation}
where $1/\lamc_{k\ell}$ is the variance associated with component $\ell$, and
$1/\lamc_k$ is the resulting aggregate variance at that point. The corresponding
precision fields $\lamc_k$ are plotted spatially in the results section to
illustrate regions of high and low model confidence.

\section{Forward problem for synthetic brain slice geometry} \label{app:forward}
The forward problem is solved using the residual module with the chosen discretization, ensuring that discretization errors are negligible.

\textbf{Domain and discretization.} The computational domain spans $\Omega = [0, 76] \times [0, 108]$ mm$^2$. The forward problem uses a rectangular mesh with $64 \times 64$ vertices, which the inverse problem also employs. Material fields use discontinuous Galerkin elements (DG0), while displacement fields use continuous linear elements (P1). For generating the displacements via the forward model, we employ the same discretization as in the inverse problem, thereby deliberately committing an 'inverse crime' \cite{kaipio2006statistical}, which ensures that discretization error remains small compared to model error. In practical applications, where this assumption does not hold, this limitation can be systematically addressed by refining the mesh density. 

\textbf{Material parameters.} The four regions are characterized as seen in Figure \ref{fig:brain_slice} and Table \ref{tab:material_properties}. The tissue density is $\rho = 1000$ kg/m$^3$.

\textbf{Harmonic excitation.} Time-harmonic loading at angular frequency $\omega = 314.16$ rad/s (50 Hz) is applied through Neumann boundary conditions $\bs{f} = [0.0, 0.1]^T$ N/mm$^2$ on the top boundary. The bottom boundary is constrained in the $s_2$-direction ($u_2 = 0$) while remaining free in the $s_1$-direction, with the bottom left corner point  fully constrained ($u_1 = u_2 = 0$) to eliminate rigid body motion.

\end{document}